%% file: main.tex
\documentclass[10pt,twocolumn,letterpaper]{article}

\usepackage[pagenumbers]{iccv} %

\input{preamble}

\definecolor{iccvblue}{rgb}{0.21,0.49,0.74}
\usepackage[pagebackref,breaklinks,colorlinks,allcolors=iccvblue]{hyperref}

\def\paperID{14711} %
\def\confName{ICCV}
\def\confYear{2025}

\def\methodName{LACONIC}

\makeatletter
\let\@oldmaketitle\@maketitle
\renewcommand{\@maketitle}{\@oldmaketitle
     \vspace{-1.4em}
     \centering
     \includegraphics[width=\linewidth]{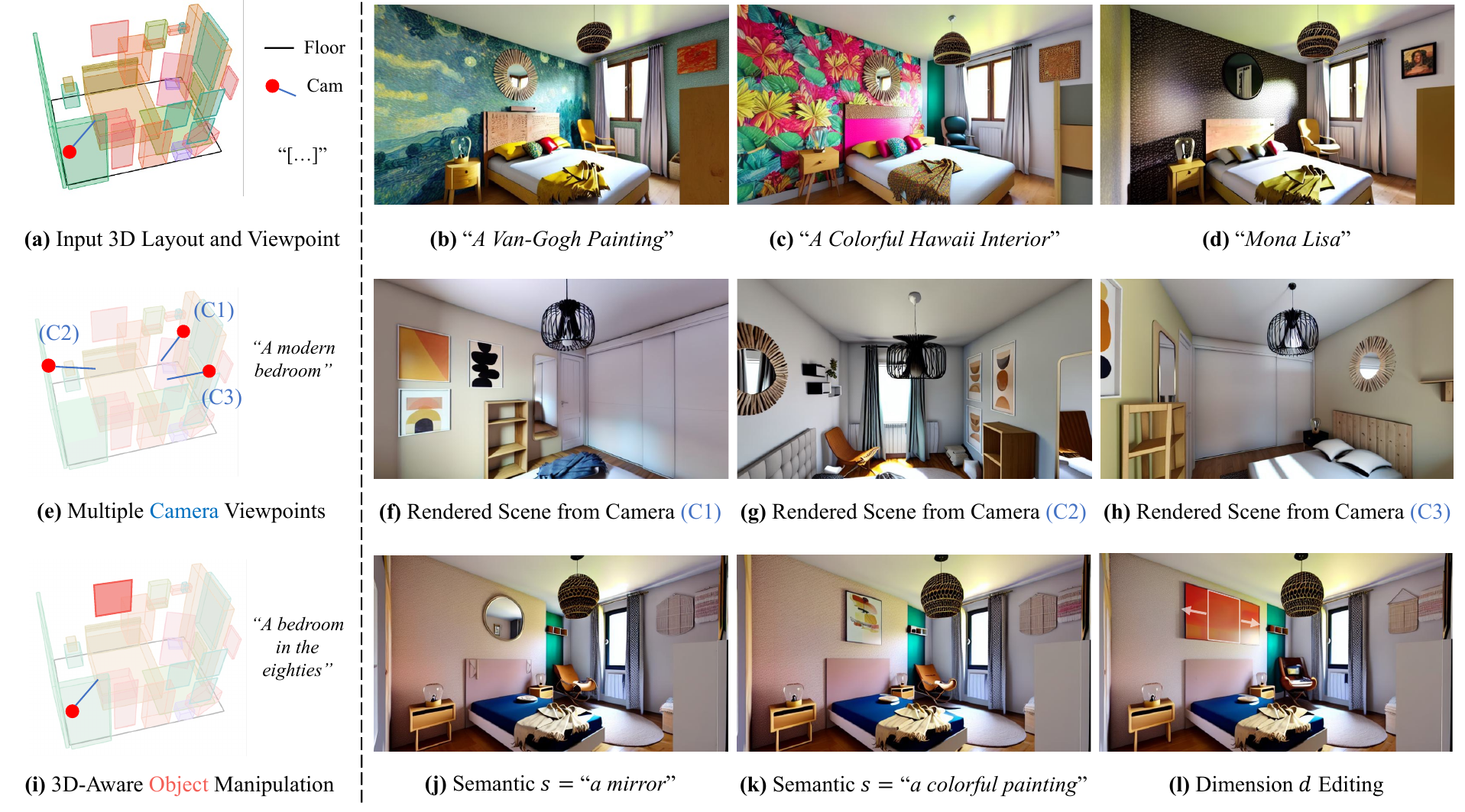}
     \captionof{figure}{\textbf{Overview of LACONIC Capabilities and Applications}. Our model generates realistic renderings from an input semantic 3D layout and target viewpoint (a), while leveraging the  comprehensive knowledge of a text-to-image prior (b--d). A given 3D scene can be rendered from multiple camera poses (e) while maintaining a consistent 3D structure across views (f--h). Finally, objects can be individually manipulated (i) by editing their open-vocabulary text caption, or position and dimension attributes in the 3D space (j--l).}
     \label{fig:teaser}
    \bigskip}
\makeatother

\title{\methodName{}:~A 3D Layout Adapter for Controllable Image Creation}

\author{Léopold~Maillard\textsuperscript{1,2}
\and
Tom~Durand\textsuperscript{2}
\and
Adrien~Ramanana Rahary\thanks{Work done during an internship at Dassault Systèmes.}
\and
Maks~Ovsjanikov\textsuperscript{1}
\and
\textsuperscript{1}LIX, École~Polytechnique, IP~Paris \qquad \textsuperscript{2}Dassault~Systèmes\\
{\tt\small \{maillard,maks\}@lix.polytechnique.fr}
\qquad
{\tt\small \{firstname.lastname\}@3ds.com}
}

\begin{document}

\maketitle

\input{sec/0_abstract}    
\input{sec/1_introduction}
\input{sec/2_related}
\input{sec/3_method}
\input{sec/4_experiments}

\input{sec/5_conclusion}

\section*{Acknowledgments}

We thank the anonymous reviewers for their insights and suggestions. We also thank Farah Ellouze and Ana Marcusanu for their feedback on the paper draft. This work was supported by Dassault Systèmes SE. The views and conclusions contained in the paper are those of the authors and should not be interpreted as representing official policies, either expressed or implied, of the company.

{
    \small
    \bibliographystyle{ieeenat_fullname}
    \bibliography{main}
}

\input{supp/X_supp.tex}
\end{document}

%% file: preamble.tex
\usepackage{multirow}
\usepackage{subcaption}
\usepackage{xcolor}
\usepackage{float}
\usepackage[accsupp]{axessibility}
\definecolor{alocasia}{RGB}{76, 164, 50}
\definecolor{bed}{RGB}{236, 184, 144}
\definecolor{wardrobe}{RGB}{148, 169, 217}
\definecolor{decorative}{RGB}{242, 162, 250}
\definecolor{window}{RGB}{192, 153, 175}

%% file: sec/0_abstract.tex
\begin{abstract}
  Existing generative approaches for guided image synthesis of multi-object scenes typically rely on 2D controls in the image or text space. As a result, these methods struggle to maintain and respect consistent three-dimensional geometric structure, underlying the scene. In this paper, we propose a novel conditioning approach, training method and adapter network that can be plugged into pretrained text-to-image diffusion models. Our approach provides a way to endow such models with 3D-awareness, while leveraging their rich prior knowledge. Our method supports camera control, conditioning on explicit 3D geometries and, for the first time, accounts for the entire context of a scene, \ie, both on and off-screen items, to synthesize plausible and semantically rich images. Despite its multi-modal nature, our model is lightweight, requires a reasonable number of data for supervised learning and shows remarkable generalization power. We also introduce methods for intuitive and consistent image editing and restyling, e.g., by positioning, rotating or resizing individual objects in a scene. Our method integrates well within various image creation workflows and enables a richer set of applications compared to previous approaches.
\end{abstract}

%% file: sec/1_introduction.tex
\section{Introduction}
\label{sec:introduction}
The integration of user-prompted conditioning signals from diverse  modalities has been a major factor in the recent success of image generation models~\cite{li2019controllable,rombach2022high,nichol2022glide,imagen}, enabling precise control, expanding creative possibilities, and narrowing the gap between human intent and generated content. Yet, there is still room to make controllable image synthesis of multi-object scenes more intuitive and better integrated with conventional design and composition workflows. In a standard pipeline for photorealistic rendering, designers generally follow a bottom-up approach: first defining the three-dimensional structure of the environment, then introducing objects and their appearance before rendering from different viewpoints and, optionally, refining details through local edits or relighting. However, existing guiding mechanisms for generative image models predominantly rely on text~\cite{rombach2022high,nichol2022glide,imagen} or image~\cite{rombach2022high,t2iadapter,controlnet,ipadapter} representations, which do not inherently preserve consistent, unambiguous, and well-defined 3D structures.

The dominant approach of generative image modeling through text guidance, while convenient, also makes it difficult to convey complex compositional structures~\cite{park2021benchmark}, as it can be cumbersome to capture nuanced spatial and geometric relations through text alone. Such relations are especially prominent in multi-object environments such as indoor scenes. Consequently, approaches leveraging structural controls, such as bounding box layouts, keypoints or semantic maps~\cite{gligen,rombach2022high,phung2024grounded}, have emerged to provide more explicit control over composition. Key to their adoption has been their efficient integration into pretrained diffusion models via the training of lightweight \textit{adapter} modules~\cite{controlnet,t2iadapter,ipadapter}, that preserve prior generative capabilities, while enabling richer and more intuitive controls. Attempts aimed at providing 3D-aware control for image synthesis and editing have gained popularity~\cite{buildascene,scenecraft,diffusionhandles,controlroom3d,ctrlroom,neuralassets}, but typically rely on panorama images, video datasets or conditioning through depth maps that are hard to acquire and offer limited flexibility.

Overall, and as illustrated in Figure~\ref{fig:overall}, the nature of these inputs is inherently associated with critical limitations, as they discard crucial spatial information such as viewpoint, object orientations, or occlusions in nested arrangements. As a result, existing representations struggle to accurately account for cases where objects are placed within others—such as books stored inside a shelf—despite such configurations being highly prevalent in real-world scenes. They also encode geometric information in a highly viewpoint-dependent manner, causing inconsistencies in how spatial structures are perceived across views. Finally, conditioning should ideally encode 3D context in a comprehensive manner. This means, for instance, capturing the influence of off-screen elements, such as lighting from a window outside the frame, but also maintaining stylistic and functional coherence across the entire scene.

In this context, we propose to inject simple, explicit 3D geometric information as guidance for single-view generative models. Associated challenges are numerous in light of the limited availability of image data with accurate 3D semantic annotations~\cite{hypersim,3dfront,scannet++}, and the lack of established solution to encode 3D inputs as additional control to a pretrained diffusion backbone. From a technical perspective, incorporating conditioning in a form that is different from text or image inputs is indeed highly non-trivial. As a result, our work aims to introduce the representational and architectural components as well as the training dynamics to \textit{augment} pretrained text-to-image models with guidance capabilities from parametric 3D semantic layouts. More precisely, our main contributions are threefold:

\begin{enumerate}
  \item A parametric conditioning representation, based on semantic bounding boxes, which, for the first time, allows for 3D-informed image synthesis and editing, maintaining consistent structures across views and without relying on depth estimators, multi-view or panorama images.
  \item An adapter architecture that establishes a tight relation between inputs in the 3D domain and the image target, while being compatible with pre-existing conditioning modalities.
  \item An efficient training framework that enables camera control as well as geometric and free-form semantic object-level guidance in 3D.
\end{enumerate}

We demonstrate the effectiveness of our method in a wide range of experimental settings, highlighting multiple advantages over current approaches. We additionally introduce a novel evaluation methodology that allows to evaluate the adherence to object-level conditioning, and use it to assess different image synthesis methods.

%% file: sec/2_related.tex
\section{Related Work}
\label{sec:related_work}

\begin{figure*}[t]
  \centering
  \begin{subfigure}[b]{0.24\textwidth}
    \scriptsize \textit{``A square bedroom with a \textcolor{bed}{king-size bed}, flanked by two nightstands. A \textcolor{wardrobe}{wardrobe}, approximately the same width as the bed, stands in front of it, with \textcolor{decorative}{decorative items} inside. Next to one of the nightstands, there is a large \textcolor{alocasia}{Alocasia plant}. A \textcolor{window}{window} is positioned in front of the plant and beside the wardrobe. The plant is at the forefront, partially obstructing the view.''}
    \vspace{3pt}
      \vfill
      \vspace{4pt}
      \caption{Text Description}
      \label{subfig:text}
  \end{subfigure}%
  \hfill
  \begin{subfigure}[b]{0.24\textwidth}
      \centering
      \includegraphics[width=\textwidth]{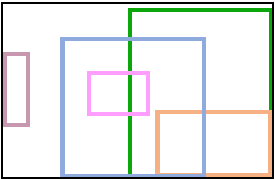}
      \vfill
      \vspace{5pt}
      \caption{Typed 2D Bounding Boxes}
      \label{subfig:image1}
  \end{subfigure}%
  \hfill
  \begin{subfigure}[b]{0.24\textwidth}
      \centering
      \includegraphics[width=\textwidth]{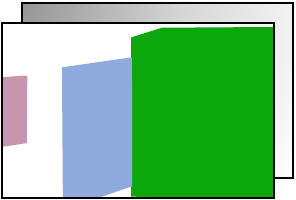} 
      \vfill
      \vspace{4.6pt}
      \caption{Semantic \& Depth Maps}
      \label{subfig:image2}
  \end{subfigure}%
  \hfill
  \begin{subfigure}[b]{0.24\textwidth}
      \centering
      \includegraphics[width=\textwidth]{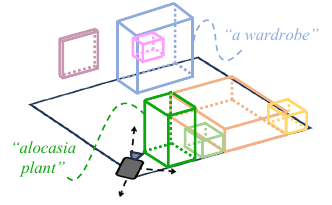}
      \vfill
      \vspace{2pt}
      \caption{\textbf{Ours}}
      \label{subfig:image3}
  \end{subfigure}
  \caption{\textbf{Comparison of high-level conditioning input representations for describing a 3D scene}. Relying solely on text descriptions (a) can make it difficult to convey complex spatial relations. Conditioning via 2D bounding boxes (b) can lead to ambiguity in perspective and does not account for out-of-bound objects. Semantic and depth maps rendered from 3D bounding boxes (c), as introduced in recent work~\cite{scenecraft}, cannot always handle occluded and nested items while being limited to objects typed from a fixed set of categories. In contrast, our layout representation encodes a comprehensive 3D scene structure that is consistent across camera views and supports object-level semantic captioning and direct manipulation of position and orientation in 3D space.\vspace{-2mm}}
  \label{fig:overall}
  \end{figure*}

\paragraph{3D-Aware Content Creation}

Recent advances in deep generative modeling~\cite{ddpm,song2021scorebased,karras2022elucidating,dit,lipman2023flow} have facilitated the emergence of powerful methods for user-driven content synthesis. In particular, score-based models have been employed to create realistic images~\cite{rombach2022high,imagen,nichol2022glide}, videos~\cite{makeavideo,ho2022video,moviegen} or 3D assets~\cite{12345,singapo} of unprecedented quality. Despite these capabilities, achieving \textit{3D-aware} image synthesis and editing, \ie, that naturally incorporate the three-dimensional structure of the underlying scene, remains challenging. This task is typically tackled by leveraging multi-view~\cite{watson2023novel,3danovelview,gu2023nerfdiff,gao2024catd,mvdiffusion} or multi-frame~\cite{neuralassets,michel2024object} image datasets, or by using depth maps, which are hard to acquire and manipulate in real-life settings, as additional conditioning inputs~\cite{3da2ddm,diffusionhandles,3deditingdepth}. 3D-aware approaches have notably demonstrated advantages over those relying on traditional text-to-image models for 3D scene generation~\cite{text2room,ctrlroom,controlroom3d}. However, these typically rely on costly score distillation~\cite{poole2023dreamfusion} from the 2D prior~\cite{mvdream,setthescene,scenecraft}, limiting their applicability.

\paragraph{Layout-Guided Image Synthesis}

A prolific line of work has been incorporating structured spatial information to control image generation. Notably, \textit{compositional} synthesis can be achieved from the guidance of 2D semantic bounding boxes~\cite{gligen,rombach2022high,yang2023reco,zheng2023layoutdiffusion,trainingfreecrossattention} or segmentation maps~\cite{rombach2022high,wang2022semantic}. Closely related to our work, SceneCraft2D~\cite{scenecraft} renders 3D indoor layouts from any camera viewpoint by projecting object bounding boxes to depth and semantic maps, that are used as conditions to train respective ControlNet~\cite{controlnet} modules. Other methods like ControlRoom3D~\cite{controlroom3d} or Ctrl-Room~\cite{ctrlroom} further rely on panorama images to render 360° consistent views of rooms. Recently, Build-A-Scene~\cite{buildascene} proposes a training-free approach based on attention guidance~\cite{trainingfreecrossattention} to perform 3D layout control from a depth-conditioned prior~\cite{loosecontrol}. While convenient, this method does not scale to \textit{complex} 3D layouts featuring more than a few objects. One common limitation of these approaches is that their conditioning input ultimately lies in the 2D space, making it difficult to disentangle objects in complex arrangements or to account for unseen \textit{contextual} items from the 3D environment.

\vspace{-2mm}
\paragraph{Adapters for Diffusion Priors}

\textit{Adapters} are lightweight learnable modules to be plugged into pretrained diffusion models. Notably, \textit{low-rank} adapters~\cite{lora} have been widely used~\cite{stylus} to efficiently perform customized domain adaptation, while \textit{structure} adapters~\cite{controlnet,t2iadapter} have enabled to control image generation with various 2D conditions that are spatially aligned with the content of the target image. More related to our work in its methodology, IP-Adapter~\cite{ipadapter} augments text-to-image models with image guidance, allowing to finely control the style and appearance of the synthesized content. This is done by learning additional cross-attention~\cite{vaswani2017attention} weights, linking image embeddings extracted by pretrained foundation models~\cite{clip,dinov2} with visual features from the diffusion backbone. In contrast, we propose to learn our 3D semantic layout encoder and attention-based adaptation module jointly.

\paragraph{Encoding Semantic 3D Layouts}

Previous lines of work have proposed a range of representations based on graphs~\cite{commonscenes,para2021generative,li2019grains} to embed 3D arrangements in a way that can capture the relations between interacting elements. Concurrently, methods defining layouts as unordered sequences of objects, to be encoded by attention-based architectures, have shown compelling results in the context of 3D scene synthesis~\cite{atiss,legonet,diffuscene,debara}.\\\\
Our method broadens the scope of existing control mechanisms and is the first to enable comprehensive control of both object semantics but also the underlying 3D geometric structure in a single unified framework.

%% file: sec/3_method.tex
\section{Method}

\begin{figure*}
\includegraphics[width=\textwidth]{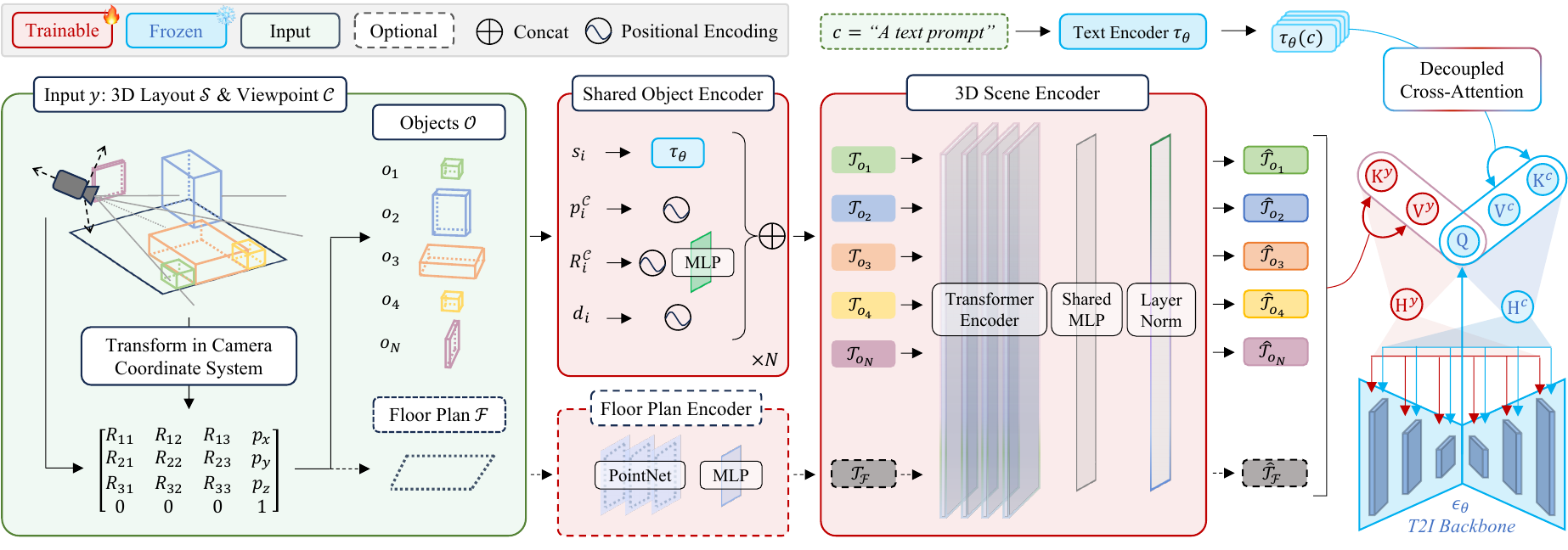}
\caption{\textbf{LACONIC Architecture \& Pipeline Overview}. From an input 3D layout \(\mathcal{S}\) and camera pose \(\mathcal{C}\), trainable modules embeds geometric and semantic properties of individual objects to guide a text-to-image diffusion prior in denoising a target rendering. Camera control is enabled by expressing spatial features \((p,R)\) from the input object 3D bounding boxes \(\mathcal{O}\) in the coordinate system defined by the target viewpoint. Resulting objects and, optionally, a room's floor plan \(\mathcal{F}\), are embedded by dedicated modules, yielding a set of representations \(\mathcal{T}\) processed by a transformer-based encoder. Output sequence of embeddings is subsequently integrated as additional conditioning input to the pretrained T2I backbone via decoupled cross-attention~\cite{ipadapter}.\vspace{-2mm}
}
\label{fig:pipeline}
\end{figure*}

In this section, we introduce our framework, which is summarized in Figure~\ref{fig:pipeline}. Our ultimate goal is to provide 3D layout control to pretrained text-to-image (T2I) models. Unlike previous work~\cite{scenecraft}, we use an explicit, view-independent 3D representation and jointly learn feature extractors and adapter modules in a single training experiment.

\subsection{Preliminaries}
\label{sec:preliminaries}

Our framework leverages text-to-image diffusion models (DM). We introduce the key associated generative and conditioning mechanisms that are relevant for our approach, and which we build upon below.

\vspace{-2mm}
\paragraph{T2I Diffusion Models}

T2I diffusion models are trained to denoise data samples \(x\), perturbed by Gaussian noise \(\epsilon\) across multiple timesteps \(t\), given their associated text caption \(c\) and, optionally, a task-specific condition \(y\). This is done by parameterizing a neural network \(\epsilon_{\theta}\) to predict the noise residual in \(x_t\) following the simple objective: 
\begin{equation}
  \label{eq:loss}
  \mathcal{L}_{\text{DM}} = \mathbb{E}_{x,c,y,\epsilon\sim\mathcal{N}(0,I),t} \Bigl[ \lVert \epsilon - \epsilon_{\theta}(x_t,t,c,y) \rVert_{2}^{2} \Bigr].
\end{equation}
Central to our approach, this learning objective can be used to incorporate diverse conditioning controls \(y\) through fine-tuning experiments, leveraging a pretrained, frozen T2I backbone alongside additional trainable components.

\vspace{-2mm}
\paragraph{Cross-Attention Conditioning}

Input signals to control text-to-image generative models are learned through the use of cross-attention layers within the architecture. We introduce here the established mechanisms and notations that intertwine with our own conditioning approach, described in Section~\ref{sec:architecture}. In particular, given a tokenized text caption \(c=[c_1, c_2, \ldots, c_M]\), a pretrained foundation model like CLIP~\cite{clip} computes a sequence representation \(\tau_\theta(c) \in \mathbb{R}^{M\times d_\tau}\) where \(d_\tau\) is the text embedding size. Resulting \textit{key} \(K^c\) and \textit{value} \(V^c\), of hidden dimension \(d\), are obtained from respective learnable linear projections. Similarly, the intermediate \(h\times w\) feature map with \(d_{\text{img}}\) channels from the image backbone, that is typically implemented as a UNet~\cite{unet} with residual blocks, is flattened and projected to the \textit{query} \(Q\). New hidden state \(H^c \in \mathbb{R}^{hw\times d}\) incorporating the text condition \(c\) is finally computed via dot-product attention~\cite{vaswani2017attention}. In practice, this mechanism is applied in a multi-headed fashion and at various feature resolutions within the conditional image denoiser \(\epsilon_\theta\).

\subsection{Semantic Layout Representation}
\label{sec:representation}

At the heart of our method lies the additional conditioning input \(y\) that we define as an intuitive and explicit proxy for furnished 3D layouts. As described in this section, individual objects are encoded from their high-level semantic and spatial properties using a parametric bounding box representation in the 3D space, as described below.

\vspace{-2mm}
\paragraph{3D Scene Parameterization}

We represent a 3D scene \(\mathcal{S}\) as an unordered set of \(N\) objects \(\mathcal{O} = \{o_1, \ldots, o_N\}\) and, optionally, an indoor floor plan modeled by a point cloud \(\mathcal{F} \in \mathbb{R}^{P\times 3}\), where \(P\) points are sampled along its boundary contours. We make \(\mathcal{F}\) optional because (i) 3D layout datasets often lack detailed structural data, like floors, ceilings and walls, and (ii) we empirically found that 3D objects alone can implicitly reflect the scene’s structure via their spatial arrangements, given that they are commonly positioned on floors or aligned against walls. We represent each object by its ``semantic 3D bounding box'': \(o_i = (p_i, d_i, R_i, s_i)\). Here \(p_i \in \mathbb{R}^{3}\) is the object's center position in the world coordinate system,  \(d_i \in \mathbb{R}^{3}\) is the size along each dimension, and \(R_i \in \mathbb{R}^{3 \times 3}\) is the rotation matrix. Similar to the global scene caption \(c\), the object-level semantic description \(s_i= [s_i^1, s_i^2, \ldots, s_i^M]\) is processed to \(M\) tokens from natural language. Those design choices are motivated by the recent success of similar lightweight representations, which have demonstrated their expressiveness in related 3D scene generative tasks~\cite{atiss,debara,diffuscene,legonet}.

\vspace{-2mm}
\paragraph{Camera Viewpoint}

The camera \(\mathcal{C}\) from which to render the 3D layout is represented by its extrinsic parameters, \ie, position \(p_\mathcal{C} \in \mathbb{R}^{3}\) and rotation \(R_\mathcal{C} \in \mathbb{R}^{3 \times 3}\). Note that, for the purpose of our general methodology, we assume that the image samples \(x\) from the data distribution are rendered using consistent camera intrinsics.

\subsection{Adapter Architecture}
\label{sec:architecture}

In this section, we introduce the trainable modules and mechanisms designed to (i) capture meaningful representations from the scene conditioning input and (ii) incorporate them to the T2I backbone, which, as described in Section~\ref{sec:preliminaries}, is conditioned through cross-attention modules and remains frozen to fully retain its original capabilities.
	
\paragraph{3D Layout Encoder}
The part of our architecture that is responsible for encoding the input 3D layout representation is inspired by previous work~\cite{legonet,debara,atiss}. Notably, individual objects \(o_i\) are handled by a common module that embeds scalar spatial features through sinusoidal positional encoding~\cite{vaswani2017attention} and dense layers, while encoding the semantic caption using the pretrained text encoder backbone \(\tau_\theta\). Concatenation of the resulting attributes produces per-object tokens \(\mathcal{T}_{o_i}\).
The optional indoor floor plan is encoded by a PointNet~\cite{pointnet} module, leading to an individual token \(\mathcal{T}_{\mathcal{F}}\). The sequence defined by the embedded tokens is passed to a transformer encoder computing new representations. Subsequently and following previous work~\cite{ipadapter}, a dense unit shared between tokens outputs representations of dimension \(d_{\tau}\), which is consistent with the global text embeddings, and is followed by Layer Normalization~\cite{layernorm}.

\paragraph{Decoupled Cross-Attention}

\begin{figure*}[t]
  \includegraphics[width=\textwidth]{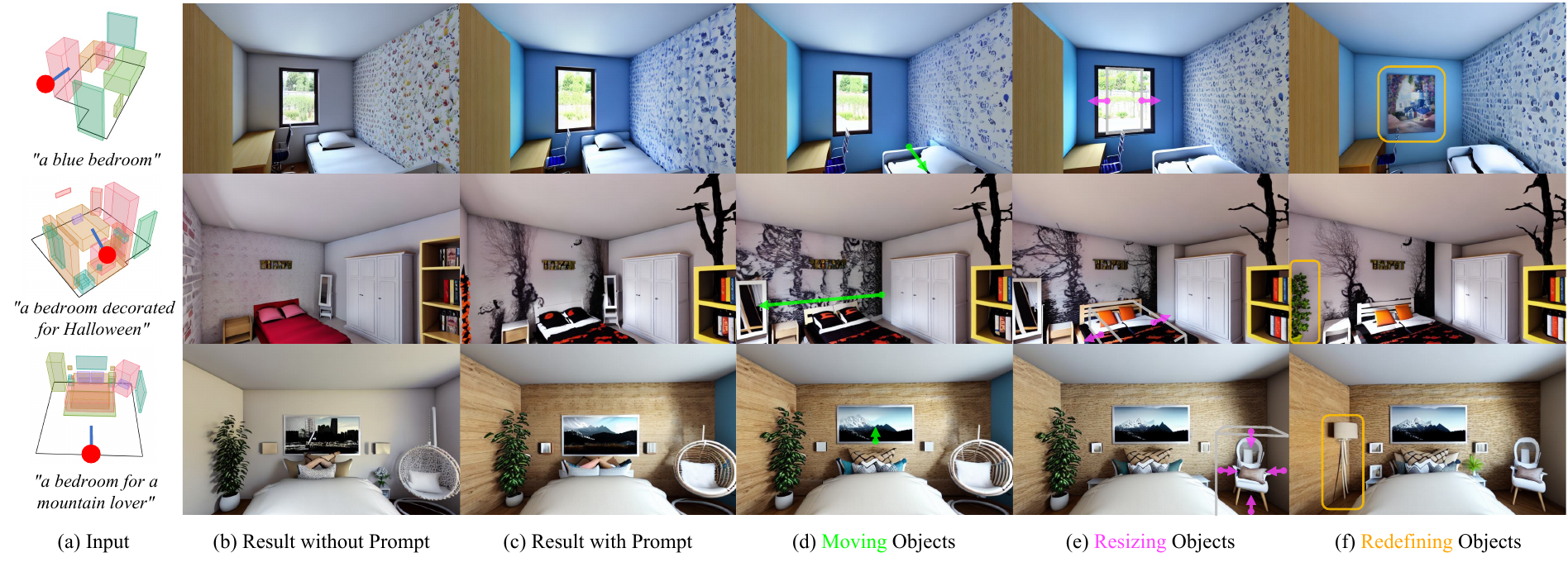}
  \caption{\textbf{Iterative Scene Editing Results}. From left to right: given an input semantic 3D layout and camera viewpoint (a), we render the scene both without (b) and with (c) global text prompt conditioning. Then, individual objects are subsequently moved (d), resized (e) and re-captioned (f). Results highlight the ability of our approach to additively perform local manipulations and generalize to out-of-distribution concepts through the use of the global text conditioning, while maintaining a consistent 3D structure. Remarkably, we observe that semantic concepts from the global text prompt are represented on \textit{relevant} objects in the generated image, \eg, patterns on wallpapers or bedsheets, art in wall frames, and do not \textit{leak} to \eg, ceilings or floors. Interestingly, removing a window has an impact on the global illumination of the scene, which would have not been possible by editing local objects via \eg, 2D image inpainting~\cite{repaint}.
  }
  \label{fig:quali-custom}
  \vspace{-5pt}
\end{figure*}

The unordered sequence of scene representations \(\hat{\mathcal{T}}=\{\hat{\mathcal{T}}_\mathcal{F}, \hat{\mathcal{T}}_{o_i},\ldots,\hat{\mathcal{T}}_{o_N}\}\) output by the 3D layout encoder is intuitively ideal to establish the individual visual contributions of each object within the image backbone through cross-attention conditioning. It is projected accordingly to \textit{key} \(K^y\) and \textit{value} \(V^y\) both in \(\mathbb{R}^{(N+1)\times d}\) by additional trainable linear projections. The hidden state associated to the input 3D layout is computed similarly to the one \(H^c\) from the text condition, via dot-product attention with the same \textit{query} \(Q\) obtained from image feature maps:
\begin{equation}
  \label{eq:scene-ca}
  H^y = \text{softmax}\Biggl(\frac{Q {(K^y)}^\top }{\sqrt{d}}\Biggr) \cdot V^y, \quad H^y \in \mathbb{R}^{hw\times d}
\end{equation}
Following the decoupled cross-attention methodology~\cite{ipadapter}, the final hidden state incorporating both the global text \(c\) and 3D layout \(y\) conditions is obtained via a weighted sum:
\begin{equation}
  \label{eq:dca}
  H = H^c + \gamma H^y, \quad H \in \mathbb{R}^{hw\times d}
\end{equation}
where \(\gamma\) is a scalar controlling the strength of the scene control with respect to the global caption one. It is finally projected back to be added to image feature maps of the diffusion backbone's residual blocks.

\subsection{Supervised Training}
\label{sec:training}

At training time, we assume to be given a collection of scenes, from which to extract structural and semantic 3D bounding box representations, as described in Section~\ref{sec:representation}. Each scene \(\mathcal{S}\) is associated to a rendering image \(x_0\) and corresponding camera pose \(\mathcal{C}\). In this section, we describe the training dynamics adopted to efficiently map our additional guidance signal to the target scene image, enabling free camera control and object-level, open-vocabulary semantic conditioning. 

\vspace{-1mm}
\paragraph{Camera Viewpoint Transformation}

One important challenge associated to our method is that  our conditioning input is defined in a three-dimensional world coordinate system, while the generation target lies in the image domain. As a result, we introduce an explicit mechanism to handle the crucial translation between the 3D input representation and underlying output image. Importantly, the information of the input viewpoint from which \(\mathcal{S}\) is rendered is directly integrated into the object spatial representations, by expressing them in the 3D coordinate system defined by the camera \(\mathcal{C}\). More precisely, we apply the following series of transformations:
\begin{equation}
  \label{eq:transposition}
  p_i^{\mathcal{C}} = R_\mathcal{C}^\top(p_i - p_\mathcal{C}), \quad R_i^\mathcal{C} = R_\mathcal{C}^\top R_i, \quad \forall o_i \in \mathcal{O}
\end{equation}
We found this reframing mechanism of the 3D input signal to the 2D image domain to be key in order to efficiently leverage the pretrained diffusion model. It can be performed in closed form on the fly, without requiring the network to learn such complex mapping between domains. When it is provided, a similar transformation is performed with the floor point cloud input \(\mathcal{F}\). In practice, object rotation features \(R_i^\mathcal{C}\) are further mapped to a continuous representation, following~\cite{zhou2019continuity}.
The resulting scene serves as the additional conditioning input \(y=\mathcal{S}^\mathcal{C}\) to guide the pretrained T2I model \(\epsilon_\theta\) in denoising the target image \(x_0\) consistent with the 3D layout structure and camera view, following the training objective from Equation~\eqref{eq:loss}.

\vspace{-1mm}
\paragraph{Conditioning Dynamics}

We adopt classifier-free guidance~\cite{ho2021classifierfree} training by randomly dropping the 3D layout input \(y\), which, at each iteration, is set to \(\emptyset\) with probability \(\mathbf{p}_{\text{drop}}\), so that the denoiser network models both the layout-conditional and unconditional image densities. Additionally, image targets \(x\) do not need to be associated to a global textual description and, as a result, \(c\) is always obtained from an empty prompt during training. Unlike previous work~\cite{scenecraft}, which relies on \textit{one-hot} category representations, our method conditions individual objects with free-form text captions \(s_i\). If not directly part of the dataset annotations, caption supervision can be derived from a vision-language model~\cite{llava,li2022blip} applied to an image showcasing the object of interest.

\subsection{Application Scenarios}

As illustrated in Figure~\ref{fig:teaser}, we provide a broad overview of the range of generative and editing capabilities enabled by our trained model at test time. Here, we consider an initial input scene \(\mathcal{S}_0\), which can be conveniently user-provided by disposing, sizing and describing a collection of objects.

\paragraph{Generating Structurally-Consistent Views}

Although our method is designed to be trained on single-view datasets, the comprehensive 3D context provided by our input representation allows synthesizing multiple views of a given scene, sharing a consistent 3D structure. More formally, from a set of target camera viewpoints \(\{\mathcal{C}_1, \ldots, \mathcal{C}_C\}\), the corresponding inputs \(\mathcal{S}_0^{\mathcal{C}_i}\) conditioning the trained model can be obtained by applying the space transformation logic from Equation~\eqref{eq:transposition}.

\paragraph{Text-Driven Scene Restyling}

Remarkably, while our model has not been trained with global text caption \(c\) supervision, its adapter design allows to fully leverage the prior knowledge from the T2I backbone, benefiting from strong out-of-distribution generalization to a wide range of semantic concepts. Notably, as described in Section~\ref{sec:additional-results}, strict adherence to the input scene and viewpoint can be relaxed by lowering \(\gamma\) in Equation~\eqref{eq:dca}, thus interpolating structural variations matching the global input description.

\paragraph{Object Attribute-Level Scene Editing}

The initial input scene \(\mathcal{S}_0\) can be iteratively edited based on user preferences, by modifying parts of the objects in \(\mathcal{O}_0\) or adjusting its items set through additions or removals. Importantly, individual objects can be manipulated with per-attribute granularity \ie, by individually and independently adjusting their size, positioning or semantic features. This shows practical advantages over previous approaches~\cite{diffusionhandles} whose attempts to solve 3D-aware image editing required to edit conditioning depth maps in the pixel space.

%% file: sec/4_experiments.tex
\section{Experiments}

\begin{figure*}
  \includegraphics[width=\textwidth]{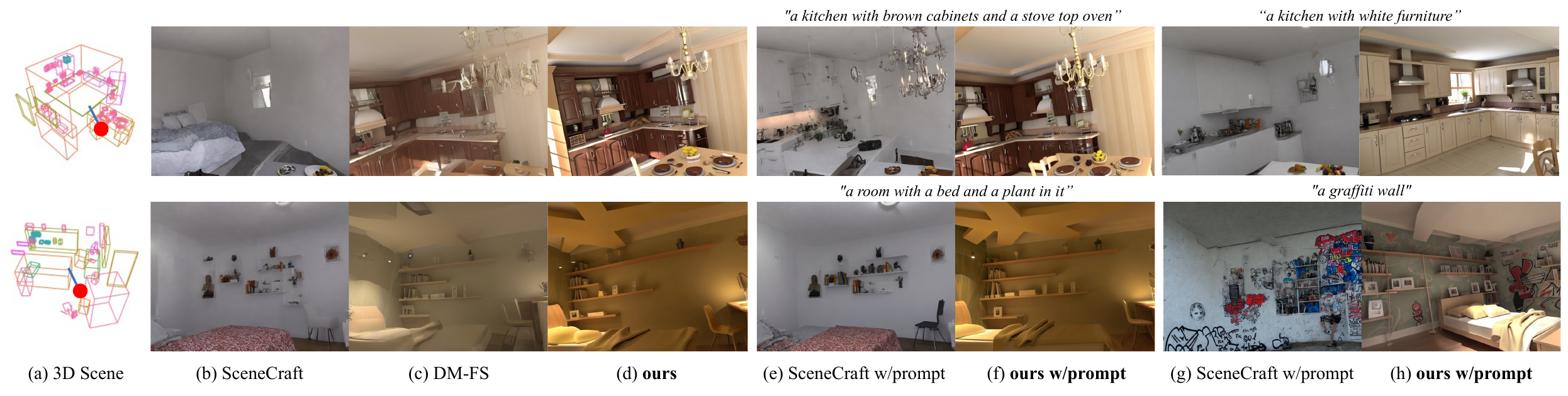}
  \caption{\textbf{3D layout-guided image synthesis baseline comparisons}. Our method produces more detailed and natural images compared to baseline approaches. Methods leveraging our 3D layout encoder (DM-FS \& ours) better represent the guiding layout, while our adapter-based approach additionally shows advantages at producing higher quality images. We also observe in text-driven scenarios (e--h) that compared to SceneCraft~\cite{scenecraft}, LACONIC better blends conditioning inputs coming from different sources, adhering both to elements from the 3D layout and the semantic text caption conditionings.
  }
  \label{fig:quali-hypersim}
  \end{figure*}

In this section, we showcase our method's capabilities at generating realistic images that incorporate the desired semantic and geometric input information. To this end, we conduct various experiments comparing it to baseline approaches on a range of standard metrics. We also introduce a novel metric that captures the alignment between the 3D layout prompt and the rendered image on the object level.

\subsection{3D Layout-Guided Image Synthesis}
\label{sec:experiments-3Dis}

\paragraph{Datasets}

We follow previous work~\cite{scenecraft} and use HyperSim~\cite{hypersim} indoor scene dataset to extract semantic 3D bounding box layouts as well as camera poses with their associated photorealistic renderings. As HyperSim only features \(326\) unique 3D layouts, paired to a total of \(24{,}383\) images, methods taking as input the comprehensive 3D layout associated to each image sample, such as ours, may tend to memorize individual fixed scenes from the dataset, making it difficult to perform the complex out-of-distribution manipulations of individual objects that our framework enables. In response, and in order to provide additional qualitative results showcasing the full range of our model's capabilities, we also gather a custom dataset featuring \(72{,}000\) \textit{bedroom} 3D scene layouts, each being paired to a single, high-quality image and viewpoint. See the Supplementary Material Section~\ref{sec:supp-dataset} for additional details and statistics on the used datasets.

\vspace{-2mm}
\paragraph{Baselines}

We ensure a relevant and fair comparison by competing against SceneCraft~\cite{scenecraft}, a recent baseline which, in the like of our method, (i) proposes to synthesize 2D images from 3D layout controls, (ii) is based on supervised training on single-view, non-panoramic images and (iii) is an adapter-based approach, enabling text-driven synthesis at test time. Additionally, we evaluate the impact of using our adapter approach instead of training a layout-conditioned \textit{Diffusion Model From Scratch} (DM-FS) on layout-image pairs by introducing a dedicated baseline that retains the 3D layout conditioning encoder from our methodology.

\vspace{-2mm}
\paragraph{Implementation}

\begin{figure*}
  \includegraphics[width=\textwidth]{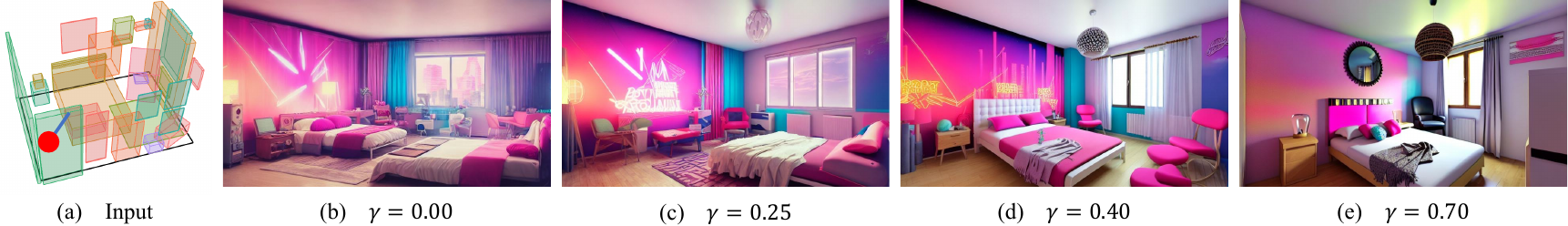}
  \caption{\textbf{Impact of the adapter strength on generated images.} From a text caption \(c=\) \textit{``A Retro Synthwave Indoor Bedroom''}, adherence with the input 3D layout improves accordingly with the parameter \(\gamma\) from Equation~\eqref{eq:dca}, achieving strict control at higher scales. We can observe that even at lower scales (c) the control enforces the geometric and semantic plausibility of the synthesized image.
  }
  \label{fig:ip-scale}
  \vspace{-4pt}
\end{figure*}

Throughout our experiments, we employ Stable Diffusion v1.5~\cite{rombach2022high} as the pretrained text-to-image diffusion backbone upon which our adapter network is trained. We train the DM-FS baseline using a similar but downsized UNet architecture, to account for the scale of the dataset. SceneCraft~\cite{scenecraft} is used with its pretrained weights and default parameters from the official implementation. All the compared methods in quantitative evaluations and side-by-side qualitative comparisons are trained on the same HyperSim~\cite{hypersim} subset. We provide comprehensive implementation details, training and inference settings, and an ablation study in the supplementary material.

\vspace{-2mm}
\paragraph{Metrics}
We evaluate generation quality and diversity with respect to the image data distribution by reporting the Fréchet Inception Distance (FID)~\cite{fid} and Kernel Inception Distance (KID \(\times 1{,}000\))~\cite{KID}. Rendering realism and variety is further assessed using the Inception Score (IS)~\cite{is}. For methods supporting text-driven synthesis, we additionally report the CLIP Score (CS)~\cite{clipscore} as a measure of how well the synthesized images align with a global caption describing the target scene. 

\paragraph{Scene Object CLIP score (SOC)}
We also evaluate different methods on a new metric that we call Scene Object CLIP score (SOC). Our score is motivated by the fact that most established metrics for evaluating image generative models primarily focus on image quality, diversity, and prompt adherence, rather than spatial or semantic alignment at the object level. Instead, SOC explicitly aims at measuring the alignment of individual objects in the conditioning layout with the corresponding locations in the synthesized image.

\begin{figure}[H]
  \includegraphics[width=\linewidth]{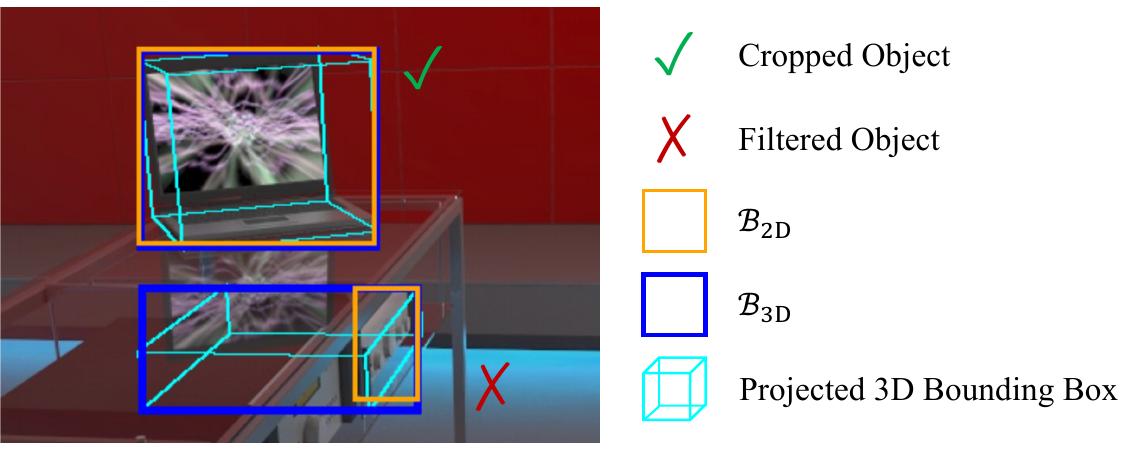}
  \caption{\textbf{Object selection for the SOC metric}. Obstructed and out-of-bounds objects are identified from their 2D and 3D bounding box annotations and filtered out from the evaluation set.
  }
  \label{fig:metric}
\end{figure}
Our metric takes as input a synthesized image, a guiding 3D semantic layout and 2D object bounding box \(\mathcal{B}_{\text{2D}}\) annotations from the ground truth image. To compute SOC, we proceed as follows: for each object in the 3D layout, we (i) project its 3D bounding box on the rendered image using the camera \(\mathcal{C}\) extrinsic and intrinsic parameters, (ii) derive the corresponding 2D enclosing bounding box \(\mathcal{B}_{\text{3D}}\), where \(\mathcal{B}_{\text{2D}} \subset \mathcal{B}_{\text{3D}}\) and (iii) define a minimum threshold on the intersection area ratio \(\mathcal{B}_{\text{2D}} / \mathcal{B}_{\text{3D}}\) for an object to be considered sufficiently visible. Finally, we crop the selected items from their enclosing bounding box \(\mathcal{B}_{\text{3D}}\), obtaining a collection of per-object images \(x_{o_i}\). These steps are meant to only keep objects in 3D space which are expected to be sufficiently visible in the given 2D image, as illustrated in Figure~\ref{fig:metric}.

Since each visible object \(o_i\) is associated with a text description, we can compute a semantic correlation between the object's text caption \(s_i\) and its crop image \(x_{o_i}\). For this last step, we follow \cite{clipscore} among others, and use the pretrained CLIP~\cite{clip} model. We provide details on our Scene Object Clip score (SOC) in the supplementary Section~\ref{sec:soc}.

\begin{table}[h]
  \renewcommand*{\arraystretch}{1.0}
     \centering
     \caption{\textbf{Quantitative experiment results on 3D layout-guided images synthesis.} We compare our method against other learning-based approaches that we outperform on most evaluation metrics, both with and without providing a global text caption \(c\).}
     \resizebox{\linewidth}{!}{%
     \begin{tabular}{@{\hskip 1mm}c c c c c c c@{\hskip 1mm}}
         \toprule
                  \multicolumn{2}{c}{Methods}& FID~$\downarrow$     & KID $\downarrow$   & IS $\uparrow$   & CS $\uparrow$  & SOC $\uparrow$ \\
         \midrule
                  SceneCraft~\small\cite{scenecraft} &\multirow{3}{*}{\small\textit{\shortstack{w/o\\text prompt}}} & 39.36 & 28.26 & 7.72 & --- & 17.59 \\
                  DM-FS & & 15.83 & 7.29 & 8.69 & --- & 18.22 \\
                  \textbf{ours} & & \textbf{9.50} & \textbf{3.44} & \textbf{9.74} & --- & \textbf{18.36} \\
          \midrule
                  SceneCraft\small~\cite{scenecraft} &\multirow{2}{*}{\small\textit{\shortstack{w/\\text prompt}}} & 27.69 & 15.21 & \textbf{14.55} & \textbf{19.75} & 17.40 \\
                  \textbf{ours} & & \textbf{10.12} & \textbf{3.91} & 10.60 & 19.74 & \textbf{18.39} \\
         \bottomrule
     \end{tabular}
     }
     \vspace{-5pt}
     \label{tab:quantitative}
 \end{table}

 \paragraph{Results} Our main quantitative results are summarized in Table~\ref{tab:quantitative}. We report results both with and without text prompting. In the latter case our approach achieves state-of-the-art results across all metrics and outperforms the very recent baseline SceneCraft \cite{scenecraft} by a significant margin. When using text conditioning (the captions provided in HyperSim \cite{hypersim}), our method still outperforms SceneCraft \cite{scenecraft} on most metrics and in several cases by a significant margin. We note that our approach is able to capture the  HyperSim data distribution significantly better compared to SceneCraft~\cite{scenecraft}. However, due to the relative limited scope of 3D structures in that dataset, our approach demonstrates somewhat less variety in text-driven scenarios as indicated by a lower Inception Score. Importantly, our adapter-based approach outperforms the DM-FS approach on all metrics, while enabling text-conditioned synthesis, thus validating the importance of using a rich pretrained T2I model. We provide qualitative comparisons on the 3D layout-guided image synthesis task in Figure~\ref{fig:quali-hypersim}. Notably, we observe that our approach produces more natural images, featuring fewer visual artifacts compared to baseline methods, while being more faithful to the input 3D layout and viewpoint. These qualitative findings are confirmed by a perceptual study detailed in the Supplementary Material Section~\ref{sec:perceptual}.

\subsection{Additional Qualitative Results}

We provide additional generation results, showcasing the intuitive and iterative editing capabilities of our framework as well as its ability to leverage the pretrained text-to-image backbone for out-of-distribution generalization in Figure~\ref{fig:teaser} and Figure~\ref{fig:quali-custom}. Overall, these results highlight that our model can capture nuanced individual object relations within structured scenes. Moreover, our approach enables manipulations in 3D space, and provides wholistic scene modeling, accounting, \eg, for off-screen objects.  Finally, results reported in Figure~\ref{fig:ip-scale} show how introducing our adapter control, even at low scale \(\gamma\), allows synthesizing 3D-aware images that are more geometrically and semantically sound than base text-to-image outputs, while maintaining adherence with the textual description. Additional qualitative results can be found in the supplementary, including a comparison against Build-A-Scene~\cite{buildascene} in Section~\ref{sec:bas} and, most notably, a demonstration of our adapter's compatibility with a modern DiT backbone~\cite{sd3} in Section~\ref{sec:dit}.
\label{sec:additional-results}

\vspace{-1pt}

%% file: sec/5_conclusion.tex
\section{Conclusion, Limitations \& Future Work}

In this paper, we introduced LACONIC, a novel 3D layout-conditional image generative framework that is distinct in both representation and architecture, expanding the creative capabilities in 3D-aware image synthesis and editing. By efficiently mapping the underlying 3D representations to the image domain and integrating them to pretrained text-to-image model, our method introduces fine-grained structural and semantic control, enhancing both user interactivity and the expressiveness and coherence of generated content.

While powerful and capable of generalization to unseen concepts, our model's output domain is still bounded by the training data distribution. As a result, a model trained on bedroom layout data will be unlikely to synthesize \eg, plausible kitchens while allowing fine-grained control. We believe that our model would be ideal to derive continuous 3D scene representations, such as NeRFs, that finely adhere to user-defined structures from the conditioning input. We also believe that exploring ways to couple geometric and \textit{visual} consistency of objects across changes in camera views is an important future research direction.

%% file: supp/X_supp.tex
\appendix

\twocolumn[{%
\renewcommand\twocolumn[1][]{#1}%
\maketitlesupplementary
\begin{center}
    \centering
    \captionsetup{type=figure}
    \includegraphics[width=\textwidth]{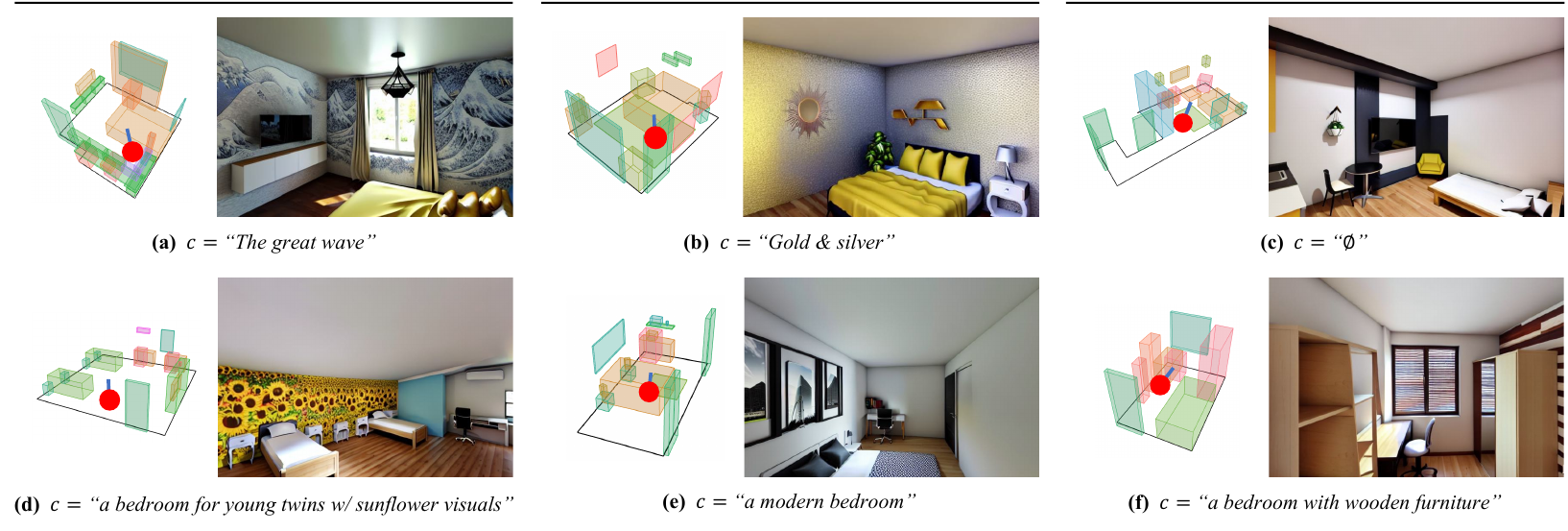}
    \captionof{figure}{\textbf{Additional layout-guided image synthesis results from diverse prompts}. The generated images (Right) from the input 3D layouts (Left) and text prompts demonstrate our method's strong adherence to both conditions. Notably, LACONIC produces high-quality results across various settings: without a text prompt (c), with in-domain prompts (e, f), and with out-of-distribution prompts (a, b, d).
    }
    \vspace{1em}
    \label{fig:supp-synthesis}
\end{center}%
}]

\section{Implementation}

\label{sec:supp-implementation}

In this section, we provide comprehensive implementation details of our 3D layout adapter architecture, pretrained text-to-image diffusion backbone, and our training and inference settings. Unless otherwise specified, the same settings are used for all datasets in our experiments. We also detail the training and test configurations of the baselines.

\subsection{Network Architecture}

\subsubsection{Pretrained Models}

We employ the established and widely adopted Stable Diffusion~\cite{rombach2022high} v1.5 to implement the conditional UNet image denoiser \(\epsilon_\theta\) and CLIP~\cite{clip} text prompt encoder \(\tau_\theta\), using the pretrained \href{https://huggingface.co/stable-diffusion-v1-5/stable-diffusion-v1-5}{weights and implementation} from the HuggingFace Diffusers~\cite{diffusers} library. In our framework, the text encoder is used to encode both the \textit{global} image caption \(c\) and the \textit{object-level} semantic descriptions \(s\).

\subsubsection{3D Layout Encoder}
\label{sec:layout-encoder}

We detail the modules used to embed the input semantic 3D layout \(\mathcal{S}\), which build upon previous work~\cite{legonet,debara,atiss} and are illustrated in Figure~\ref{fig:pipeline} and described in Section~\ref{sec:architecture} in the main paper.

\paragraph{Shared Object Encoder}

Parametric object bounding box representations \(o_i\) are embedded by a common module. More precisely, each scalar defining individual object spatial attributes \((p_i, R_i, d_i)\) is projected to a higher-dimensional vector using fixed, sinusoidal positional encoding~\cite{vaswani2017attention} with \(32\) frequencies. We follow~\cite{legonet} and apply:
\begin{equation}
  \label{eq:pe}
  PE(k) = \{\sin(128^{j/31}k), \cos(128^{j/31}k) \}_{j=0}^{31} \in \mathbb{R}^{64}
\end{equation}
Consequently, object's \textit{position} \(p_i\) and \textit{dimension} \(d_i\) are encoded to \(192\)-dimensional attributes. Importantly, the \textit{rotation} matrix \(R_i\) is first expressed in a continuous representation, following the recommendations from~\cite{zhou2019continuity}, leading to a vector in \(\mathbb{R}^6\) from which individual scalars are also processed by Equation~\eqref{eq:pe}. This produces a representation in \(\mathbb{R}^{384}\) that is subsequently mapped by a linear layer to \(\mathbb{R}^{192}\). The semantic embedding \(\tau(s_i)\) is the \textit{end-of-sequence} output token from the pretrained text encoder in \(\mathbb{R}^{d_\tau}\), with \(d_\tau = 768\), and is further projected to a \(192\)-dimensional representation by an MLP featuring a single hidden layer of dimension \(384\) with LeakyReLU activations. Encoded object attributes are finally concatenated, yielding an individual token \(\mathcal{T}_{o_i}\) of dimension \(4 \times 192 = d_\tau\) for each of the \(N\) objects in the scene.

\paragraph{Floor Plan Encoder}

Following prior work~\cite{legonet,debara}, we encode the optional scene's floor plan \(\mathcal{F}\) by leveraging a popular \href{https://github.com/fxia22/pointnet.pytorch}{implementation} of the PointNet~\cite{pointnet} model, applied on a point cloud representation obtained by sampling \(P=100\) three-dimensional points that are evenly spaced along the room's boundaries. A single \textit{floor} token \(\mathcal{T}_{\mathcal{F}_i} \in \mathbb{R}^{d_\tau}\) is subsequently obtained by projecting the \(1024\)-dimensional module's output.

\paragraph{Transformer Encoder}

New token representations \(\hat{\mathcal{T}}\) are established by a Transformer encoder that follows the seminal paper~\cite{vaswani2017attention} by using the implementation from the PyTorch~\cite{paszke2019pytorch} library. In practice, we use \(4\) encoder layers with \(6\) attention heads and hidden dimensions of size \(512\). Since the scene is defined as an unordered sequence of its objects and floor representations, we do not additionally encode the position of individual tokens in \(\mathcal{T}\), and perform \textit{zero-padding} to handle scenes with different number of objects \(N\). In practice, we consider a maximum number of \(50\) objects. Following~\cite{ipadapter}, the output token embeddings are each passed to a \textit{shared} MLP that preserves the token dimension \(d_\tau\), implemented with a hidden unit of size \(768\), GELU activation~\cite{gelu}, and followed by Layer Normalization~\cite{layernorm}.

\subsubsection{Cross-Attention Layers}

As introduced in Section~\ref{sec:architecture} of the main paper, the scene conditioning sequence \(\hat{\mathcal{T}}\) output by the 3D layout encoder is mapped to associated \textit{key} \(K^y\) and \textit{value} \(V^y\) by introducing respective learnable dense projection matrices \(W_K \in \mathbb{R}^{d \times d_\tau}\) and \(W_V \in \mathbb{R}^{d \times d_\tau}\). In practice, we follow the \href{https://github.com/tencent-ailab/IP-Adapter}{implementation} from IP-Adapter~\cite{ipadapter} to introduce the decoupled cross-attention mechanism within residual blocks of the pretrained Stable Diffusion UNet backbone.

\subsection{Datasets}

\label{sec:supp-dataset}

In this section, we introduce the datasets used in our experimental evaluation and describe any dataset-dependent mechanisms implemented to adapt our work to their specific features or available annotations.

\subsubsection{HyperSim Dataset}

\paragraph{Features}

Following previous work~\cite{scenecraft}, we leverage indoor scenes and photorealistic renderings from HyperSim~\cite{hypersim} to construct a collection of semantic 3D bounding box layouts with camera poses \(y\) and image \(x\) pairs. The dataset originally features \(461\) scenes from diverse types of indoor environments, and \(77{,}400\) rendered images. We follow SceneCraft~\cite{scenecraft} and leverage a filtered version of the \href{https://huggingface.co/datasets/gzzyyxy/layout_diffusion_hypersim}{dataset} proposed in~\cite{controlroom3d}, discarding unbounded and scenes with excessively large scales. This results in \(323\) unique scenes, that are associated with \(24{,}383\) images annotated by a global text caption extracted by a pretrained foundation model. Each layout features 3D bounding box annotations that are typed according to the standard NYU40~\cite{nyu40} semantic classes. Each rendering is also associated with an object instance map, that we leverage in the context of our SOC metric implementation, as detailed in Section~\ref{sec:soc}. We conducted foundational statistical analysis of the 3D layouts, establishing that scenes contain an average of \(121\) objects, with a median of \(54\) and that the dataset demonstrates a large diversity in terms of structural complexity, which is challenging in the context of our work and in light of the limited number of available samples.

\paragraph{Experimental Setup}

We describe here the dataset-specific mechanisms implemented to run experiments on HyperSim layouts. Importantly, a proportion of scenes from the dataset contain hundreds of objects, exceeding the maximum length fixed to \(50\) for our sequence-based representation used for the model conditioning. As a result, we leverage the instance semantic map of the target rendering to prioritize including visible objects in the conditioning sequence. If the number of visible objects still exceeds the limit, we identify the most meaningful objects based on their bounding box dimensions. Since the HyperSim dataset does not provide the floor information in a straightforward manner, we set \(\mathcal{F} = \emptyset\) in our experiments. Object-level semantic information are also limited to their categorical label among NYU40 classes, from which we apply \(\tau_\theta(s_i)\) as described in our general methodology.

\subsubsection{Custom Bedroom Dataset}

As described in the main paper in Section~\ref{sec:experiments-3Dis}, the limited number of available HyperSim layouts, that are passed as conditioning input, makes it easier for our model to learn a mapping between the global scenes and specific camera poses and target renderings, while overlooking the contributions of individual objects. In response, we gathered a custom dataset of \(72{,}000\) human-designed 3D indoor bedroom layouts, in which each layout is associated to a single rendering. The dataset also includes floor information, represented as a 3D point cloud. Finally, ground-truth object-level semantic descriptions \(s_i\) are extracted using a LLaVA model~\cite{llava} from their 2D rendering using the following instruction: \textit{``Describe this object concisely. Do not analyse the background, only the object''}. This level of annotations allows to implement all the building blocks from our general methodology.

\subsection{Training Protocol}

We optimize our adapter network while keeping the text-to-image denoiser frozen by following standard DDPM~\cite{ddpm} training with \(\epsilon\)-prediction objective, as formalized in Equation~(\ref{eq:loss}) in the main paper. Noise timesteps are uniformly sampled from \(t\sim\mathcal{U}(0,1000)\). At each iteration, we randomly drop the conditioning input 3D layout \(y\) with a rate \(\mathbf{p}_{\text{drop}} = 0.15\). We use the AdamW~\cite{adamw} optimizer, with weight decay coefficient \(\lambda = 0.01\) and learning rate \(\eta=5\times10^{-5}\), reached following an initial linear \textit{warmup} phase during the first \(200\) training iterations. Our checkpoints are trained for a total of \(250\) epochs, with \(\eta\) decreasing according to a cosine schedule. We leverage PyTorch Lightning~\cite{Falcon_PyTorch_Lightning_2019} to distribute experiments across \(3\) NVIDIA RTX 6000 GPUs, each handling a batch size of \(10\) samples. Importantly, we experimented with enabling \textit{automatic mixed precision} during training, but empirically observed instability and inconsistent convergence.

\subsection{Inference Settings}

At test-time, images for quantitative evaluation are generated following the sampling algorithm from DPM-Solver~\cite{dpmsolver}. We apply \(30\) denoising steps with a classifier-free guidance scale of \(5.0\). Images in our qualitative results are obtained with the DDIM sampler~\cite{ddim} using \(50\) steps.

\subsection{Baselines}

\begin{figure}
\includegraphics[width=\linewidth]{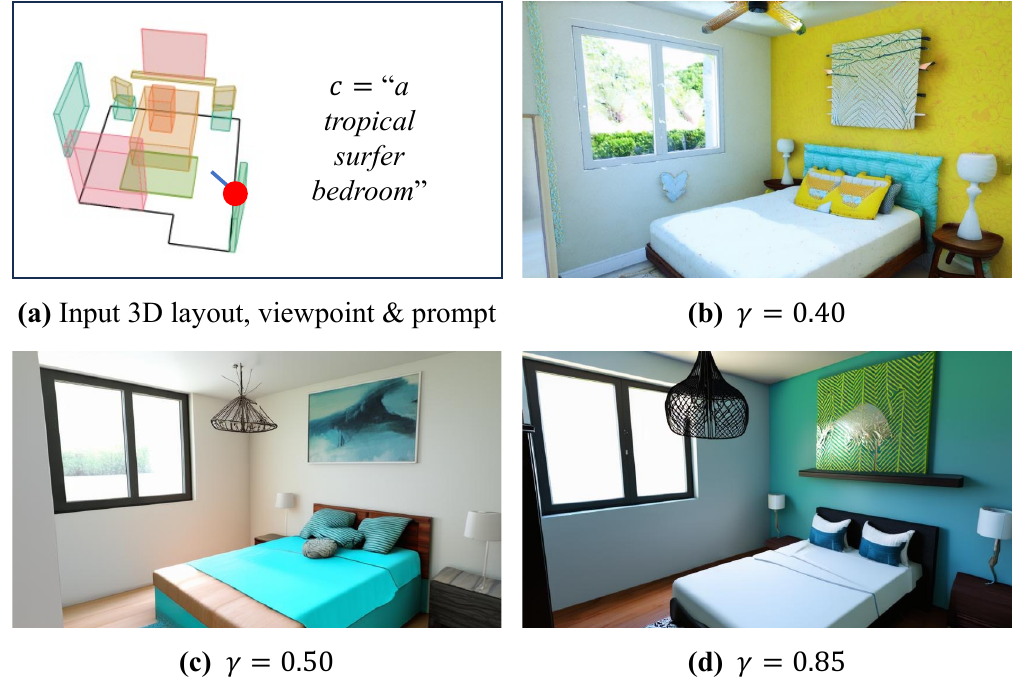}
\caption{\textbf{Text-driven DiT synthesis results}. Given an input 3D layout, viewpoint and caption (a), LACONIC with Stable Diffusion 3~\cite{sd3} supports adjusting the adapter strength to balance fidelity to the text prompt against adherence to the input layout (b--d).} 
\label{fig:dit-text-cond}
\vspace{-4pt}
\end{figure}

\subsubsection{SceneCraft}

SceneCraft~\cite{scenecraft} is a recent baseline method that proposes to tackle 3D layout-guided image synthesis, in which, similar to our approach, 3D layouts are defined as a collection of typed object 3D bounding boxes. To do so, it renders the input 3D layout from the target camera viewpoint to (i) a semantic map in which individual boxes are colored according to their one-hot semantic category and (ii) depth maps that are automatically derived from the input geometry. From this pair of 2D input representations and in order to introduce the additional controls to a pretrained text-to-image backbone, SceneCraft trains dedicated \textit{ControlNet}~\cite{controlnet} modules in a supervised experiment. In practice, we leverage the pretrained \href{https://huggingface.co/gzzyyxy/layout_diffusion_hypersim_prompt_one_hot_multi_control_bs32_epoch24}{checkpoint} obtained by training the modules on the same HyperSim~\cite{hypersim} subset used in our own experiments. It uses Stable Diffusion~\cite{rombach2022high} v2.1 as the T2I prior. We employ official author \href{https://github.com/OrangeSodahub/SceneCraft}{implementation} with the default test-time parameters.

\subsubsection{Diffusion Model from Scratch}

We describe below the main motivations and implementation details behind the \textit{Diffusion Model trained From Scratch} (DM-FS) from our experimental evaluations.

\paragraph{Intuitions}

A key design choice in our framework is to use a cross-attention-based \textit{adapter} network to augment existing T2I models with 3D layout guidance, rather than training a dedicated, conditional generative model from scratch. This choice is motivated by two primary factors. First, achieving photorealistic generation results would be extremely challenging given the small scale of available image datasets featuring 3D layout annotations, especially on less common camera views that are not widely represented in the training distribution. Second, our adapter approach allows us to leverage the powerful priors of large-scale T2I models, enabling strong generalization to other domains, as demonstrated by our experimental results. We believe that those insights validate the relevance of this baseline in the context of our contributions' evaluation.

\paragraph{Implementation}

To experimentally validate our intuitions, we trained a baseline from scratch, jointly optimizing a randomly initialized UNet and the 3D layout encoder from Section~\ref{sec:layout-encoder}. The backbone is similar to that of Stable Diffusion, implemented with Diffusers~\cite{diffusers}, and is guided through standard cross-attention conditioning~\cite{rombach2022high} between the 3D scene sequence \(\mathcal{S}\) and feature maps from the residual blocks. Due to the limited caption variety in the training data, we omit text conditioning for this baseline. To account for the smaller training set, the UNet is also downsized compared to Stable Diffusion and comprises \(4\) downsampling and upsampling residual blocks with \(128\), \(256\), \(256\) and \(384\) output channels. The two \textit{bottleneck} blocks, that operate at smaller resolutions, are augmented with cross-attention with \(8\) heads. The 3D layout encoder's architecture is identical to that described in Section~\ref{sec:layout-encoder}.

\begin{figure*}
\includegraphics[width=\textwidth]{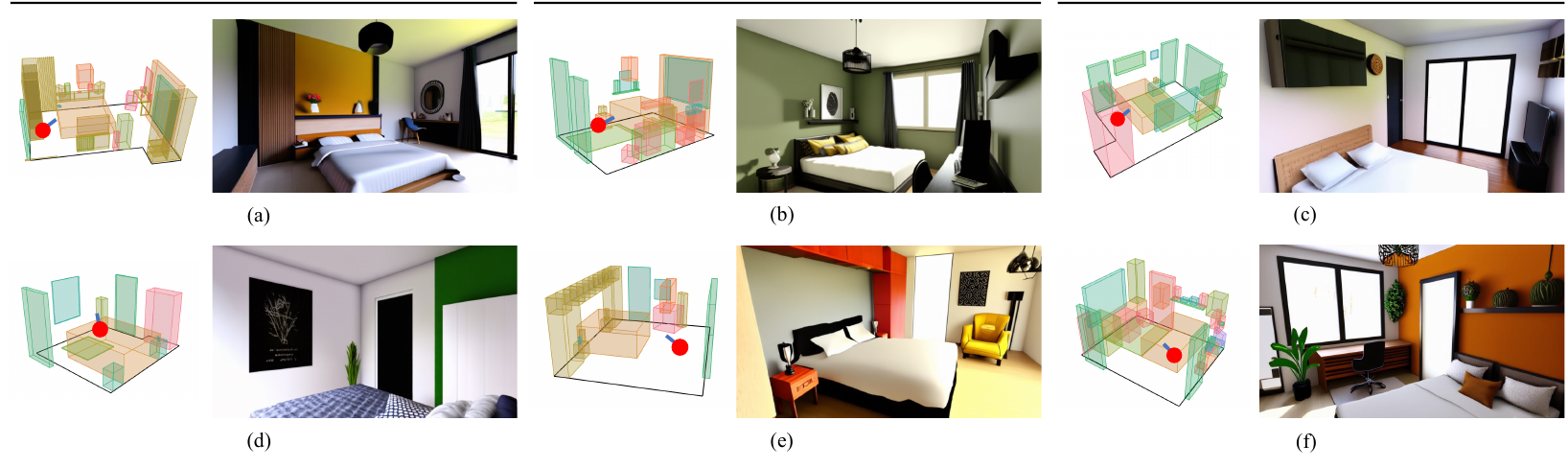}
\caption{\textbf{Layout-guided image synthesis results with a DiT-based backbone}. Our LACONIC adapter successfully conditions Stable Diffusion 3~\cite{sd3}, demonstrating compatibility with modern DiT architectures. The generated images (Right) show strong adherence to the input 3D layouts and camera viewpoints (Left).
}
\label{fig:dit-results}
\vspace{-4pt}
\end{figure*}

\subsection{Evaluation Metric}
\label{sec:soc}

In Section~\ref{sec:experiments-3Dis} of our main paper, we introduce the Scene Object CLIP (SOC) score as a robust way to simultaneously assess whether objects in the generated content (i) are correctly positioned and sized with respect to their spatial conditioning information and, at the same time, (ii) match their assigned semantic attributes.

\paragraph{Procedure}

To compute the score for a synthesized image, we first identify its main visible objects given the conditioning 3D layout \(\mathcal{S}\), the camera pose \(\mathcal{C}\), and the 2D object \(o_i\) bounding box annotations \(\mathcal{B}_{\text{2D}}^i\) from the associated ground truth image \(x_0\). Since these 2D  boxes do not assess that the corresponding objects \(o_i\) are mostly visible, we also project the 3D bounding boxes from the conditioning signal onto the image plane to derive enclosing 2D boxes \(\mathcal{B}_{\text{3D}}^i\). For each object, we notice that this second bounding box fully contains the one from the image annotation, and compute the ratio \(r_a\) as the area of \(\mathcal{B}_{\text{2D}}^i\) divided by the area of the enclosing projected box \(\mathcal{B}_{\text{3D}}^i\). Remarkably, this projection mechanism also handles objects that partially lie outside the image bounds. Based on this ratio, objects \(o_i\) that are sufficiently visible are cropped at the location of \(\mathcal{B}_{\text{3D}}^i\) in the synthesized image, yielding per-object images \(x_{o_i}\). We finally compute a CLIP~\cite{clip} correlation score between object images and their associated text semantic \(s_i\).

\paragraph{Filtering Parameters}

We rely on an object's associated visible area ratio \(r_a^i\) to determine if its crop image will be added to the evaluation base. In practice and to report the values in Table~\ref{tab:quantitative} in the main paper,  we set a threshold value \(\alpha = 0.4\). Additionally, we filter out tiny objects, which are prevalent in the HyperSim~\cite{hypersim} dataset. Indeed, objects whose instance semantic map covers less than \(2\)\% of the image area are discarded. Finally, we also filter objects whose NYU40 semantic annotation is non-descriptive, \eg, that are classified as \textit{``other''}. The final metric is averaged over all the resulting object crops in the dataset.

\section{Generalization to DiT Backbones}

\label{sec:dit}

In this section, we demonstrate the flexibility of our adapter by integrating it with a recent Diffusion Transformer (DiT)~\cite{dit} backbone. Specifically, we employ Stable Diffusion 3~\cite{sd3} to showcase that LACONIC is not limited to established UNet-based models conditioned via standard cross-attention, but is capable of remarkable generalization across architectures.

\subsection{Background on Joint-Attention}

Stable Diffusion 3~\cite{sd3} introduces a Multimodal Diffusion Transformer (MM-DiT) architecture trained with Rectified Flows~\cite{lipman2023flow}. The core of its conditioning approach is a joint-attention mechanism that operates over a unified sequence containing both image tokens (from the latent \(x_t\)) and text tokens (from the caption \(c\)). Within each transformer block, separate sets of linear projections are used to yield the respective image \(Q^x\), \(K^x\), \(V^x\) and text \(Q^{c}\), \(K^{c}\), \(V^{c}\) matrices. These are subsequently concatenated into global matrices (\eg, \(Q = [Q^x;Q^c]\)) used for attention, enabling a bidirectional information flow between text and visual features within the architecture. The resulting hidden states \(H^x\) and \(H^c\) are finally separated and processed by additional modality-specific feedforward modules.

\begin{figure*}
\includegraphics[width=\textwidth]{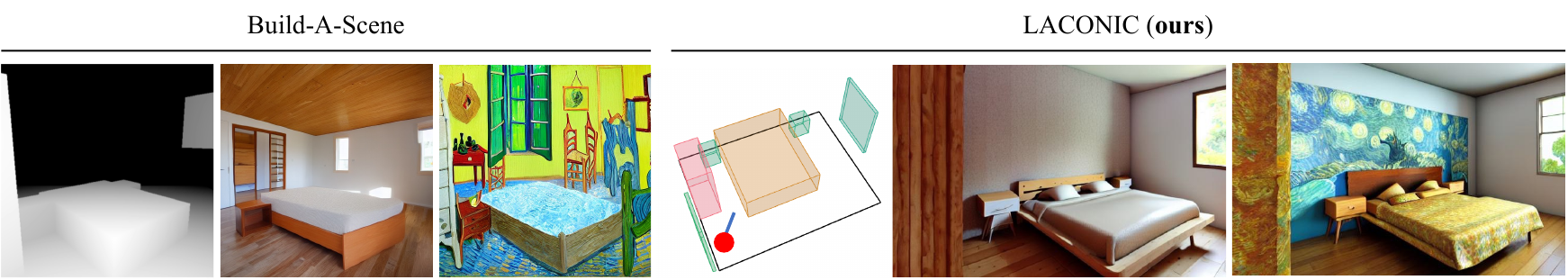}
\caption{\textbf{Qualitative comparison with Build-A-Scene~\cite{buildascene} given a common 3D layout and viewpoint}. For each method, from left to right: input scene representation, generation result for prompt \(c_1=\) ``\textit{a cozy bedroom with a wooden floor}'', and for \(c_2=\) ``\textit{a Van-Gogh style bedroom}''. Build-A-Scene's input representation also features individual prompt for each bounding box.
}
\vspace{-4pt}
\label{fig:bas}
\end{figure*}

\subsection{Layout Conditioning}

Our 3D layout condition is integrated by performing an additional joint-attention pass. Consistent with our general methodology, this is implemented by introducing new trainable linear projections to augment the pretrained model. Specifically, we obtain new key-value pairs from both the text-conditioned image state \(H^x\) and the encoded layout \(y\) features, while leveraging the query matrix \(Q\) from the primary text-conditioning step. A second joint-attention operation is then performed over concatenated matrices, producing an output that is split into a novel image state \(H^{x^\prime}\), and a layout state \(H^y\) which is processed by additional linear layers. These are merged with the outputs from the primary joint-attention pass, following the linear combination from Equation~(\ref{eq:dca}) in the main paper, \ie,
\begin{equation}
  H^{x}_{\text{final}} = H^x + \gamma H^{x^\prime} \quad \text{and} \quad H^{c}_{\text{final}} = H^c + \gamma H^y
  \label{eq:supp-strength}
\end{equation}
In practice, we leverage the Diffusers backbone \href{https://huggingface.co/stabilityai/stable-diffusion-3-medium-diffusers}{implementation and pretrained weights}, that we augment with the custom joint-attention processor \href{https://github.com/unity-research/IP-Adapter-Instruct}{implementation} from IPAdapter-Instruct~\cite{ipai}.

\subsection{Qualitative Results}

We show in Figure~\ref{fig:dit-results} images synthesized by Stable Diffusion 3~\cite{sd3} DiT after training a dedicated LACONIC adapter. Notably, the augmented backbone successfully takes into account the input 3D layout and viewpoint and, as a result, renders scenes in accordance with the specified objects spatial and semantic features. Additionally, Figure~\ref{fig:dit-text-cond} highlights that the model also supports multimodal conditioning with a non-empty global text caption \(c\), whose influence with respect to the layout can be adjusted following Equation~(\ref{eq:supp-strength}).

\section{Additional Results}

In this section, we provide additional experimental results highlighting the capabilities of our model and advantages of our design choices.

\subsection{Ablation Study}

We trained ablated versions of our model to measure the individual contribution of:

\begin{itemize}
  \item \textbf{L-R}: applying the reframing mechanism from world to camera coordinates described in Section~\ref{sec:training}.
  \item \textbf{L-T}: including the transformer encoder module in the adapter architecture detailed in Section~\ref{sec:architecture}.
\end{itemize}
Quantitative evaluation metrics are computed on the custom \textit{bedroom} dataset and reported in Table~\ref{tab:ablation}. Reported results further validate our design choices. Notably, we can observe that the transformation of the input layout to the camera's 3D coordinate system has a key contribution to our method's performance.

\begin{table}[!htbp]
     \centering
     \caption{\textbf{Ablation study on our framework's main components}.}
     \resizebox{0.65\linewidth}{!}{%
     \begin{tabular}{@{\hskip 1mm}c c c c@{\hskip 1mm}}
         \toprule
                  Ablation Setting & FID~$\downarrow$     & KID $\downarrow$  & SOC $\uparrow$ \\
          \midrule
                  L-R & 34.03 & 32.17 & 22.17 \\
                  L-T & 10.68 & 11.19 & 22.28 \\
                  \textbf{LACONIC} & \textbf{9.68} & \textbf{10.24} & \textbf{22.35} \\
         \bottomrule
     \end{tabular}
     }
     \label{tab:ablation}
     \vspace{-4pt}
 \end{table}

 \subsection{Baseline Comparison}
 \label{sec:bas}

We provide an additional qualitative comparison against the recent Build-A-Scene~\cite{buildascene} method. While fundamentally different in its setting by adopting a training-free approach, the baseline also attempts to perform 3D layout guided image synthesis. To do so, objects are iteratively added to a given background by leveraging a depth-conditioned image generative model and by manipulating the attention maps of the pretrained backbone. Qualitative results from the same input bedroom layout are reported in Figure~\ref{fig:bas}. Remarkably, our method more accurately reflects the input 3D structure, showcasing all and only the specified objects.

 \subsection{Perceptual Study}

 \label{sec:perceptual}

We conducted a perceptual study to compare our method with SceneCraft~\cite{scenecraft}, evaluating both the overall quality of the generated images and their adherence to the input 3D layout and viewpoint.

\vspace{-1mm}
\paragraph{Baseline Settings}

\begin{figure}
\includegraphics[width=\linewidth]{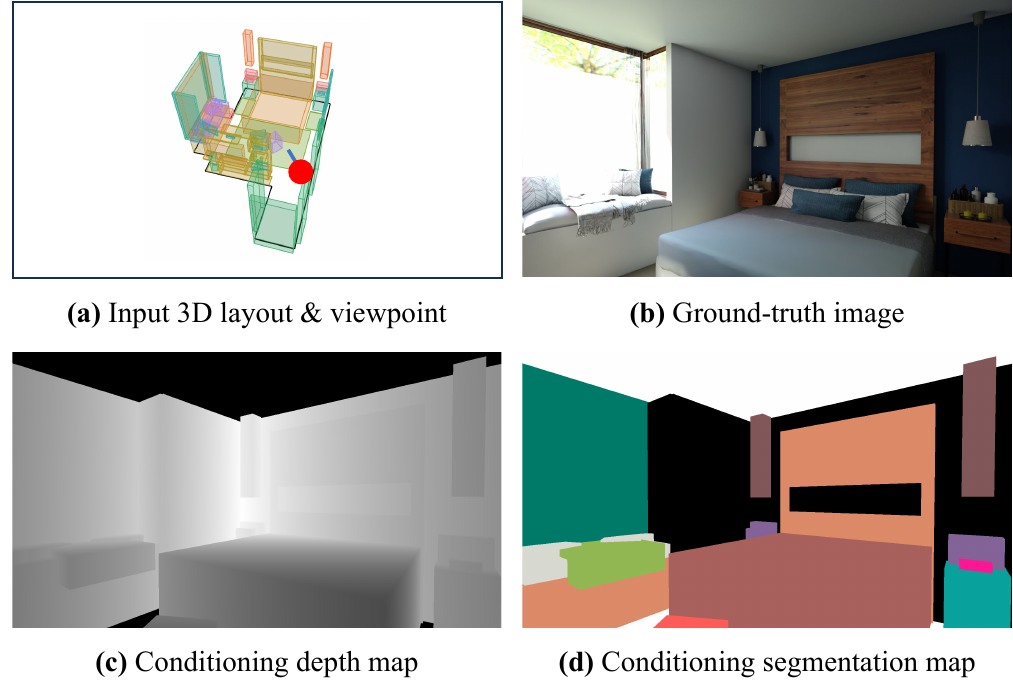}
\caption{\textbf{Conditioning inputs for the SceneCraft~\cite{scenecraft} baseline}. The model is conditioned on depth (c) and segmentation (d) maps, which are rendered from the 3D bounding box layout (a) underlying a ground-truth \textit{bedroom} image (b).} 
\label{fig:scenecraft-bedroom-cond}
\vspace{-4pt}
\end{figure}

\begin{figure*}[b]
  \centering
  \includegraphics[width=\textwidth]{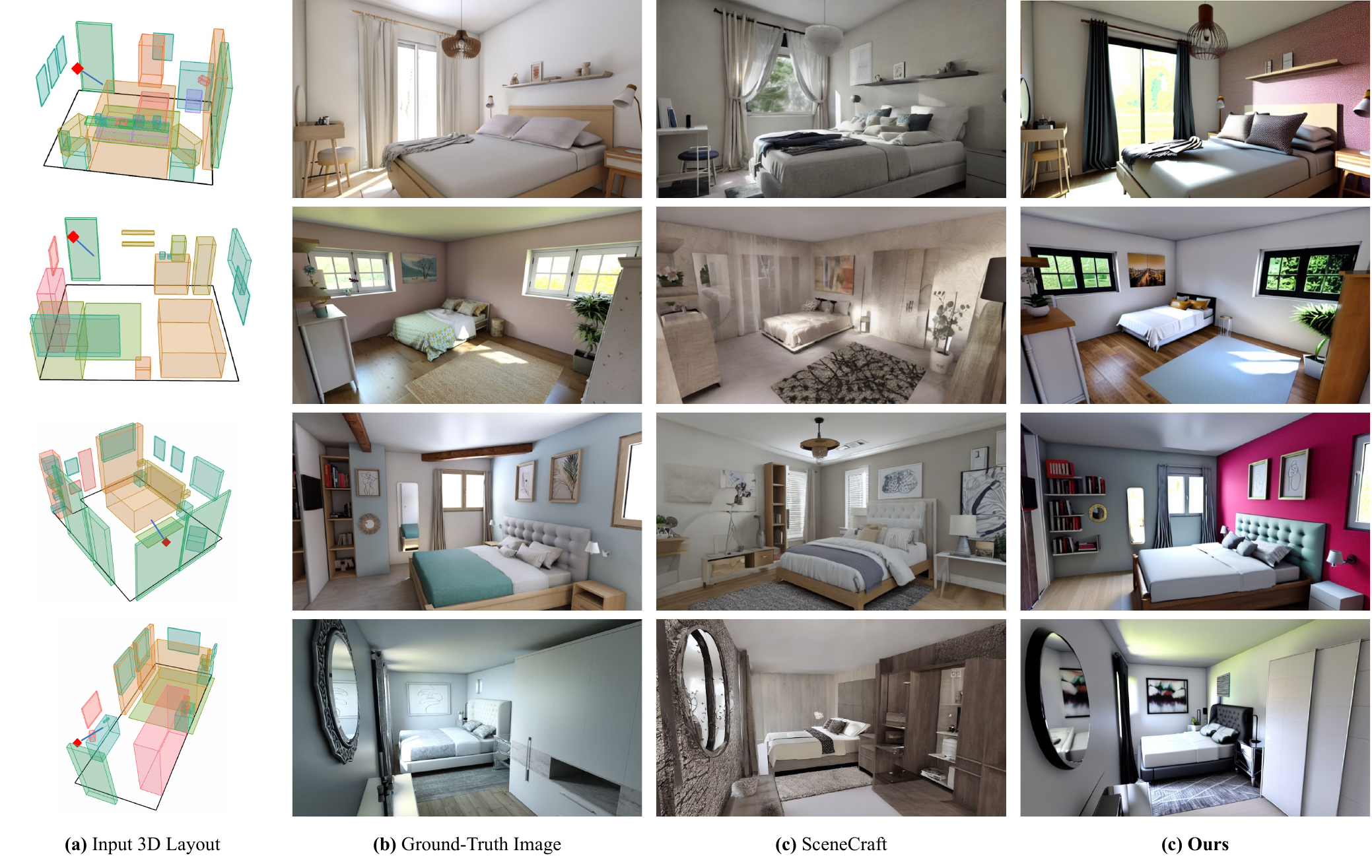}
  \caption{\textbf{Comparison with SceneCraft~\cite{scenecraft} on 3D layout-guided image synthesis.} Our method demonstrates superior realism and  adherence to the input 3D layout and viewpoint on our custom \textit{bedroom} dataset.
  }
  \label{fig:quali-bedroom}
\end{figure*}

Methods are compared on our custom \textit{bedroom} dataset on which we train a SceneCraft model by optimizing jointly respective ControlNet~\cite{controlnet} modules from depth and semantic maps. These conditioning inputs are rendered from the 3D layout and at the target camera view, as shown in Figure~\ref{fig:scenecraft-bedroom-cond}. We follow the official implementation and default parameters, using the same Stable Diffusion v1.5 backbone as our method. A qualitative comparison with this baseline is shown in Figure~\ref{fig:quali-bedroom}.

\vspace{-1mm}
\paragraph{Study Design}

\begin{figure*}[t]
  \centering
  \includegraphics[width=\textwidth]{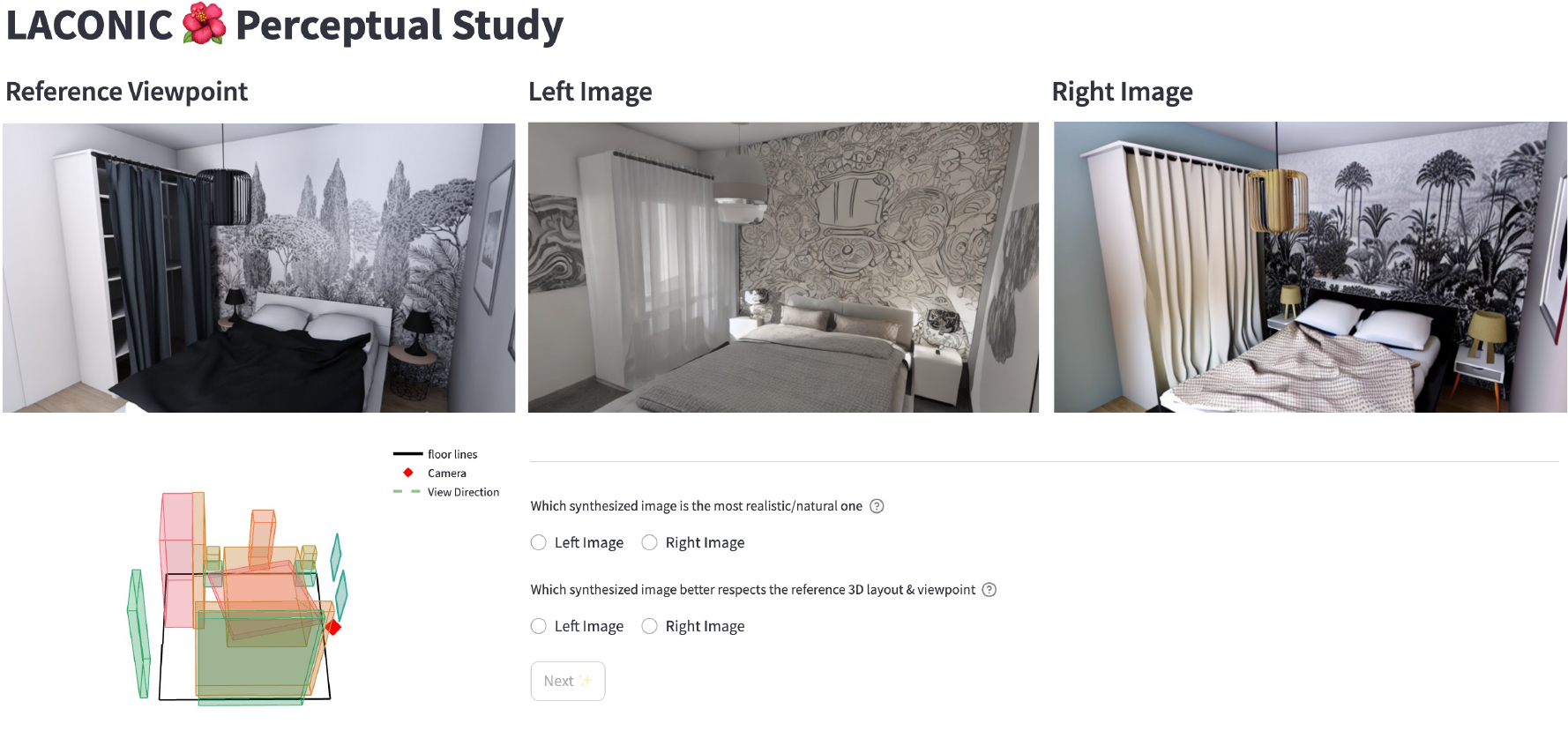}
  \caption{\textbf{Perceptual Study Interface}. Users are prompted to independently select which generation result is (i) the most realistic and (ii) more in line with the input 3D layout and viewpoint.
  }
  \label{fig:study-interface}
\end{figure*}

We employed a two-alternative forced-choice (2AFC) test. For a given 3D layout and text prompt, participants were shown the pair of synthesized images and asked to choose the better one based on two separate criteria: (i) ``Which image is the most realistic/natural?'' and (ii) ``Which image better respects the reference 3D layout and viewpoint?''. To prevent bias, the positions of the images were randomized for each trial. The study interface provided an interactive view of the 3D layout input and the ground truth image for reference, as shown in Figure~\ref{fig:study-interface}.

\vspace{-1mm}
\paragraph{Results}

We collected \(638\) preference votes from \(15\) participants across \(319\) image pairs. The results demonstrate a clear preference for our method which was favored for realism in \(71.2\%\) of comparisons. For adherence to the input 3D layout, LACONIC achieved a strong \(89.0\%\) preference rate. This trend was remarkably consistent across participants. Based on their individual average ratings, \(100\%\) of participants (\(15\) out of \(15\)) favored our method on conditioning adherence, and \(93\%\) (\(14\) out of \(15\)) favored it on realism.

\subsection{Qualitative Results}

\begin{figure*}
  \centering
  \includegraphics[width=\textwidth]{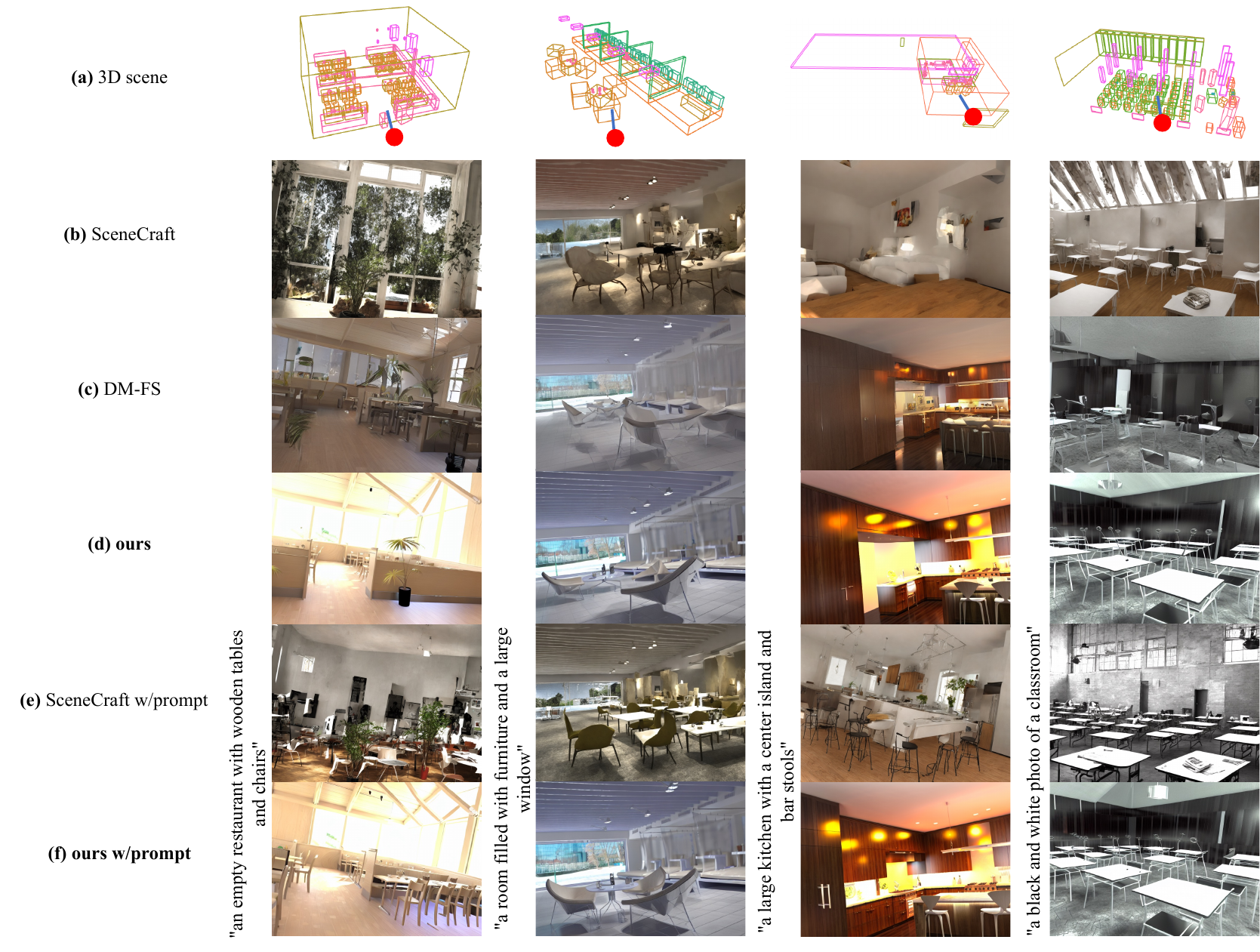}
  \caption{\textbf{Additional 3D layout-guided image synthesis baseline comparisons (1/2)}. We can observe that our method produces more natural images that better respect the input 3D layout.
  }
  \vspace{-1pt}
\label{fig:quali-hypersim-1}
\end{figure*}

\begin{figure*}
  \centering
  \includegraphics[width=\textwidth]{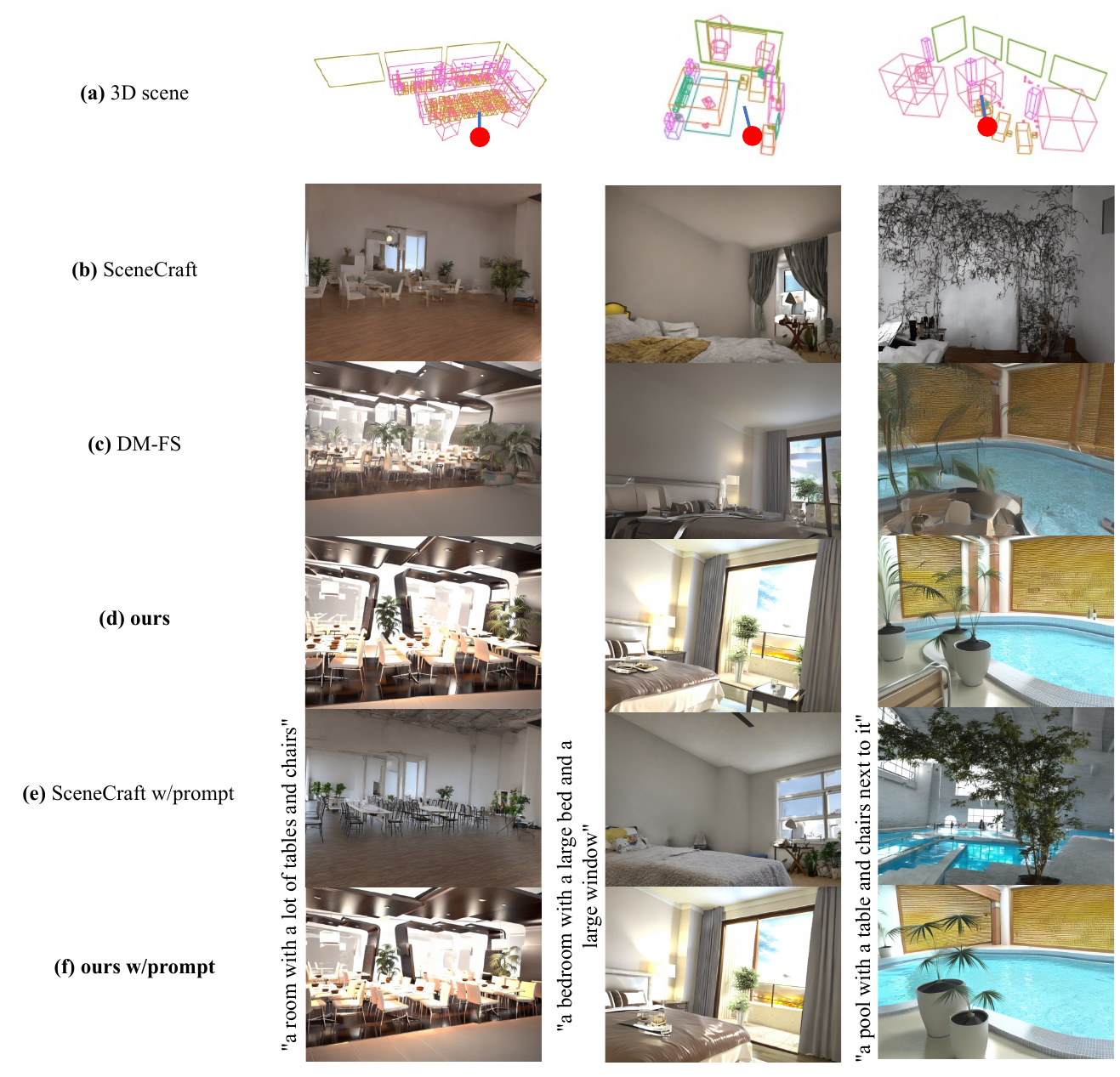}
  \caption{\textbf{Additional 3D layout-guided image synthesis baseline comparisons (2/2)}.
  }
  \vspace{-1pt}
\label{fig:quali-hypersim-2}
\end{figure*}

We provide additional qualitative results on text-driven, layout-guided image synthesis in Figure~\ref{fig:supp-synthesis}. These results highlight our method's ability to adhere to the input semantic 3D layout while achieving out-of-distribution generalization to unseen concepts. Additionally, we report in Figure~\ref{fig:quali-hypersim-1} and Figure~\ref{fig:quali-hypersim-2} supplemental comparisons against the DM-FS and SceneCraft~\cite{scenecraft} baselines, both with and without providing an additional text condition. Our method consistently generates more realistic images with superior detail, 3D layout fidelity, and text prompt adherence. Motivating our 3D layout conditioning encoder, the DM-FS baseline appears to better respect the input 3D layout in comparison to SceneCraft~\cite{scenecraft}. We notice that both adapter-based approaches, ours and SceneCraft, leverage the pretrained T2I backbone, resulting in fewer artifacts in the generated images in comparison to DM-FS. Consequently, our method demonstrates a significant advantage in complex scenes with many objects, combining the expressiveness of our 3D layout encoder with the T2I backbone’s ability to generate detailed visual features.

\section{Limitations}

\begin{figure}[b]
  \centering
  \includegraphics[width=\linewidth]{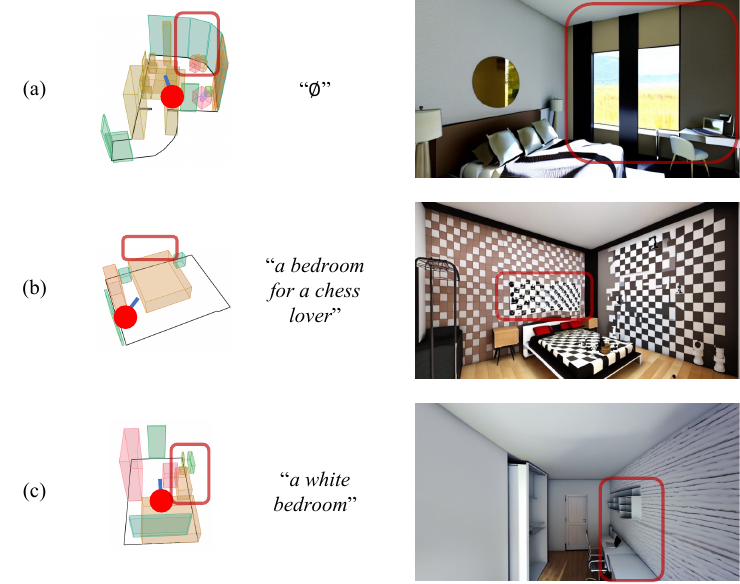}
  \caption{\textbf{Failure Cases and Limitations}. In (a), some objects, such as the windows, are missing in the generated image and the floor shape, while complex, is not accurately rendered. We can also observe visual stability issues, as shown in the result in (b), which features an inconsistent pattern on the wall resembling a misplaced frame above the bed. Finally, generated images can exhibit unnatural perspective, resulting in distorted floors and objects, as showcased in (c). This behavior may result from our assumption of consistent camera intrinsics across all data samples.
  }
  \vspace{-4pt}
\label{fig:failure}
\end{figure}

Figure~\ref{fig:failure} illustrates failure cases and known limitations. Although our method demonstrates state-of-the-art 3D layout adherence, it can occasionally generate results with inconsistencies, such as missing objects, visual artifacts, or distorted perspectives.

\section{Societal Impact}

Controllable generative AI for 3D environments has significant implications for both individuals and industries. We believe our method to predominantly yield positive societal impact. By enabling fine-grained control over generation, our approach makes 3D-aware content creation more intuitive and accessible, empowering not only artists and designers but also individuals with no prior expertise in 3D modeling. It also benefits industries that rely on realistic environmental rendering, such as urban planning, architecture, and digital twins for cities. A key advantage of our method is its parameter efficiency, which reduces the computational cost of training---an important factor given the high environmental impact of generative models~\cite{societal1, societal2}. However, these benefits come with challenges. Since our approach does not natively incorporate recent advances in generative model watermarking~\cite{societal3, societal4}, it could be misused for potentially deceptive applications. Additionally, by significantly lowering the barrier to content generation, it may contribute to rapid job displacement in traditional 3D design fields. Another concern is the potential for bias propagation, as image generative models, including semantic and style-driven ones like ours, may unintentionally reinforce stereotypes~\cite{societal5}---for example, by associating certain colors with gender, by \eg, synthesizing a bedroom as blue for \textit{``boys''} and pink for \textit{``girls''}. Ensuring the responsible development and use of such technologies, while addressing ethical concerns and mitigating biases, is crucial.

